\documentclass[letterpaper,10pt,conference]{ieeeconf}

\IEEEoverridecommandlockouts
\usepackage{amsmath}
\usepackage{amssymb}
\usepackage{graphicx}
\usepackage{booktabs}
\usepackage{makecell}
\usepackage{multirow}
\usepackage{tabularx}
\newcolumntype{C}{>{\centering\arraybackslash}X}
\usepackage{siunitx}
\usepackage[table]{xcolor}
\usepackage{xspace}
\usepackage{algorithm}
\usepackage{algorithmic}
\usepackage{wrapfig}
\usepackage{bibunits}
\usepackage{etoolbox}
\usepackage[pagebackref=true,breaklinks=true,colorlinks=true,
            citecolor=black,urlcolor=blue,linkcolor=blue]{hyperref}
\usepackage[figurelike]{contactexplorer-captions}
\providecommand{\citet}[1]{\cite{#1}}
\providecommand{\citep}[1]{\cite{#1}}

\definecolor{oursrow}{HTML}{EAF2FB}
\definecolor{stdgray}{gray}{0.55}
\newcommand{\ms}[2]{$#1{\scriptstyle\,\textcolor{stdgray}{\pm #2}}$}
\newcommand{\msb}[2]{$\mathbf{#1}{\scriptstyle\,\textcolor{stdgray}{\pm #2}}$}
\newcommand{\ourrow}{\rowcolor{oursrow}}
\newcommand{\task}[1]{\texttt{#1}}
\newcommand{\method}[1]{\textit{#1}}
\newcommand{\para}[1]{\noindent\textbf{#1.}}

\title{\LARGE \bf
ContactExplorer: Contact Coverage-Guided Exploration for General-Purpose Dexterous Manipulation}

\DeclareRobustCommand{\contactemail}[1]{{\fontfamily{pcr}\mdseries\upshape\selectfont\detokenize{#1}}}
\author{%
Zixuan Liu$^{*,1}$, Ruoyi Qiao$^{*,1}$,
Chenrui Tie$^{1}$, Xuanwei Liu$^{1}$\\[2pt]
Yunfan Lou$^{1}$, Chongkai Gao$^{1}$,
Zhixuan Xu$^{1}$, Lin Shao$^{1,2,\dagger}$%
\thanks{$^{*}$\,Equal contribution; authors listed in random order.}%
\thanks{$^{\dagger}$\,Corresponding author.}%
\thanks{$^{1}$\,School of Computing, National University of Singapore.}%
\thanks{$^{2}$\,RoboScience.}%
\thanks{\raggedright Contact: \contactemail{zixuanliu@u.nus.edu},
\contactemail{ruoyi_qiao@outlook.com}, \contactemail{linshao@nus.edu.sg}.}%
}

\newcommand{\projectlinksentence}{}
\newcommand{\teaserwidth}{0.90\linewidth}

\newcommand{\paperabstract}{%
Reinforcement learning explores effectively in domains such as Atari games, navigation, and locomotion, where novelty over states or dynamics is a sufficient signal. In contrast, dexterous manipulation requires rich physical hand--object interactions, but existing methods often suffer from unstable contact-based novelty signals, inefficient distance novelty signals, or reliance on task-specific priors.
We propose \textbf{ContactExplorer}, a general exploration method for dexterous manipulation tasks. ContactExplorer represents contact as the intersection between object surface points and hand keypoints, encouraging dexterous hands to discover diverse and novel contact patterns, namely which fingers contact which object regions. It maintains a contact counter conditioned on discretized object states obtained via learned hash codes. This counter is leveraged in two complementary ways: (1) a count-based contact coverage reward that promotes exploration of novel contact patterns, and (2) an energy-based reaching reward that guides the agent toward under-explored contact regions. We evaluate ContactExplorer on seven contact-rich manipulation tasks and five dexterous hand embodiments. Experimental results show that ContactExplorer substantially improves sample efficiency and success rates over existing exploration methods, that it reduces the need for task-specific priors, and that it remains effective across hand embodiments and transfers to the real world.}

\begin{document}

\maketitle
\thispagestyle{empty}
\pagestyle{empty}

\begin{abstract}
\paperabstract\space Website:
{\hypersetup{urlcolor=black}\fontfamily{pcr}\bfseries\selectfont
\href{https://contact-explorer.github.io/}{contact-explorer.github.io}}.
\end{abstract}


\section{Introduction}
\label{sec:intro}
Exploration has long been a central challenge in Reinforcement Learning (RL), particularly in tasks where external rewards are sparse or difficult to design. A prominent line of research addresses this challenge through intrinsic motivation, which provides auxiliary rewards for behaviors that are novel, informative, or uncertain~\cite{strehl2005mbie, strehl2008mbieeb, kolter2009BayPACOP}. These approaches have led to substantial progress in domains such as Atari games~\cite{stadie2015incentivizing,NIPS2016unify,burda2018rnd,ecoffet2019goexplore,noveld}, navigation~\cite{pathak2017icm,taiga2018stateabs,pathak2019disagreement}, and locomotion~\cite{NIPS2017hash}, where exploration can often be guided by novelty over state or dynamics.

However, dexterous manipulation presents a fundamentally different exploration challenge. Unlike locomotion tasks, which primarily focus on controlling the robot’s own body motion, manipulation requires rich physical interactions with external objects. Such interactions are realized through contact, making contact a natural basis for novelty in manipulation. Prior methods largely approximate it: hand--object distance~\cite{pmlr-v229-schwarke23a-curiosity} indicates how close the hand is, not whether contact occurs or where on the object it lands. \method{HaC}~\cite{hapticscuriosity} instead measures contact directly, using contact-force prediction error to compute novelty for manipulation with parallel grippers. Yet in dexterous manipulation, sparse and highly discontinuous contacts make force-based novelty signals noisy and unstable~\cite{jing2026contactawareneuraldynamics}. Rather than improving the novelty signal, another line of work reduces the exploration burden with task-specific priors, including carefully designed initializations~\cite{Xu_2023_unidex, good-pregrasp-init}, motion priors~\cite{chen2025object, mandi2025dexmachina}, or hand-crafted heuristics~\cite{lin2025contactmarkerrl} that guide the hand toward favorable interaction behaviors. While effective in specific settings, these methods typically rely on substantial human knowledge and lack generality across tasks and objects.

\begin{figure}[t]
  \centering
  \includegraphics[width=\teaserwidth]{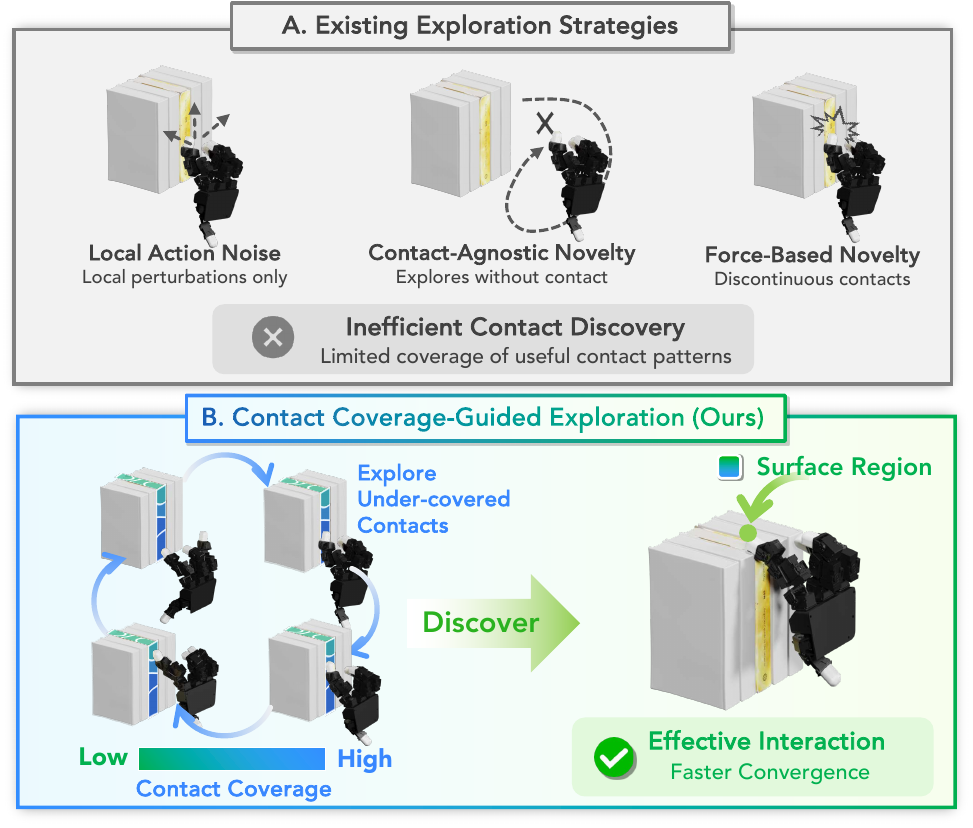}
  \caption{\textbf{Contact Exploration Strategies.} \textbf{(A)} Existing strategies discover contact inefficiently, perturbing only nearby contacts, exploring without requiring contact, or relying on forces that are unreliable under discontinuous contacts. \textbf{(B)} ContactExplorer tracks contact coverage over object surface regions and guides the fingers toward under-covered contacts, yielding effective interaction and faster convergence.}
  \label{fig:teaser}
  \vspace{-8pt}
\end{figure}

To this end, we propose \textbf{ContactExplorer}, a contact-centric exploration framework that explicitly models and incentivizes hand--object interaction on novel contact patterns, namely which fingers contact which surface regions of the object. To address the sparsity of contact, ContactExplorer combines two complementary signals: a post-contact count-based reward that encourages novel finger--region contacts and a pre-contact energy-based reaching reward that guides the hand toward regions likely to yield new interactions. Exploration is thus both contact-focused and continuously guided. To make it depend on the task phase and object configuration, ContactExplorer conditions its contact counters on both the current and goal object states.
We evaluate ContactExplorer on seven contact-rich manipulation tasks spanning single-hand and bimanual control, cluttered and constrained scenes, in-hand reorientation, and tabletop grasping (Fig.~\ref{fig:tasks}). It consistently achieves higher success rates and faster convergence than existing exploration methods, and its contact behaviors transfer to real-world systems.

In summary, this work makes three contributions. \textbf{First}, we introduce a contact representation that models hand--object interaction as finger--surface-region pairs. \textbf{Second}, we condition contact coverage on learned object-state clusters, preventing contact statistics from interfering across distinct configurations. \textbf{Third}, we derive complementary post-contact novelty and pre-contact reaching rewards, and validate their effectiveness across seven simulation tasks, five hand embodiments, and real-world deployment. These results show that structured contact coverage can improve exploration while reducing reliance on task-specific priors. \projectlinksentence


\section{Related Work}
\label{sec:related_work}

\subsection{Intrinsic Rewards for RL Exploration}
Intrinsic rewards improve exploration by encouraging novelty in the agent's experience, typically through state novelty or dynamics novelty. 
State novelty methods reward rarely visited states~\cite{strehl2005mbie, strehl2008mbieeb, kolter2009BayPACOP, burda2018rnd, NIPS2016unify, NIPS2017hash, taiga2018stateabs,ecoffet2019goexplore,noveld}. 
In dexterous manipulation, however, their effectiveness depends on the state representation: force-based signals are noisy and unstable due to force spikes and discontinuities~\cite{jing2026contactawareneuraldynamics}, while hand--object distance-based novelty~\cite{pmlr-v229-schwarke23a-curiosity} may not induce physical interaction. 
We instead measure state novelty through contact patterns, namely which fingers contact which object regions, yielding a reliable and interaction-centric exploration signal.

Dynamics novelty methods reward transitions that are difficult to predict~\cite{stadie2015incentivizing, pathak2017icm, pathak2019disagreement}. 
For manipulation, \method{HaC}~\cite{hapticscuriosity} uses contact-force prediction error to encourage contact-rich behavior in parallel grippers, but such force prediction is unreliable for dexterous manipulation due to non-smooth contacts~\cite{jing2026contactawareneuraldynamics}. 
Huang et al.~\cite{huang2019gentle} use a gentleness metric that encourages low-force interactions, but this coarse hand-level signal lacks the spatial resolution required for diverse dexterous contact strategies.

\subsection{Task Priors in RL for Dexterous Manipulation}
Beyond exploration strategies, prior work often introduces explicit task priors to improve training efficiency in dexterous manipulation. These approaches can be broadly categorized into reaching guidance and prior knowledge injection. Reaching guidance typically uses dense shaping rewards to encourage the hand to move toward the object~\cite{zhang2024graspxl, dexsingrasp, pmlr-v305-chen25b-clutterdexgrasp}. While effective for grasping-oriented tasks, such guidance is often insufficient for more complex dexterous manipulation that requires rich and diverse hand--object interactions. Other approaches incorporate prior knowledge by explicitly encoding assumptions about how the hand should interact with the object. Examples include high-quality initialization~\cite{Xu_2023_unidex, good-pregrasp-init, Wan_2023_unidex++, Khandate-RSS-23-manirrt}, manually engineered robot-centric rewards~\cite{dexpbt, chen2023sequential, bi-dex, bai2025retrieval, wang2025exdex, lin2025contactmarkerrl, dexsingrasp}, grasp generation modules~\cite{wu2023learningaffplanning, Xu_2023_unidex}, learning from human videos~\cite{objectoriented, gao2025flip}, and expert demonstrations~\cite{adroit, chen2025object, mandi2025dexmachina, showdemo}. Although effective in specific settings, these methods rely on strong task- or embodiment-specific priors, which limits their generalizability and provides limited support for structured exploration that can autonomously exploit contact for dexterous manipulation.

\section{Contact Representation and State Conditioning}
\label{sec:formulation}


\subsection{Problem Formulation}

We formulate dexterous manipulation as a Markov Decision Process (MDP), defined by the tuple
$\mathcal{M} = (\mathcal{S}, \mathcal{A}, P, r, \rho_0, \gamma)$, where $\mathcal{S}$ is the state space containing robot proprioception and object-state information, $\mathcal{A}$ is the continuous low-level action space, $P$ is the transition kernel, $\rho_0$ is the initial-state distribution, and $\gamma$ is the discount factor. The reward function $r$ combines a task-specific reward with the exploration rewards introduced in this work. We use Proximal Policy Optimization (PPO)~\cite{schulman2017ppo} to solve this MDP for all tasks. This formulation describes the training of a privileged teacher with access to object state in simulation; the real-world student is subsequently obtained by distillation and operates under partial observations (Section~\ref{sec:realworldexperiments}).

\subsection{Object and Hand Representations}

\begin{wrapfigure}{r}{0.16\textwidth}
    \centering
    \vspace{-12pt} 
    \includegraphics[width=1.0\linewidth]{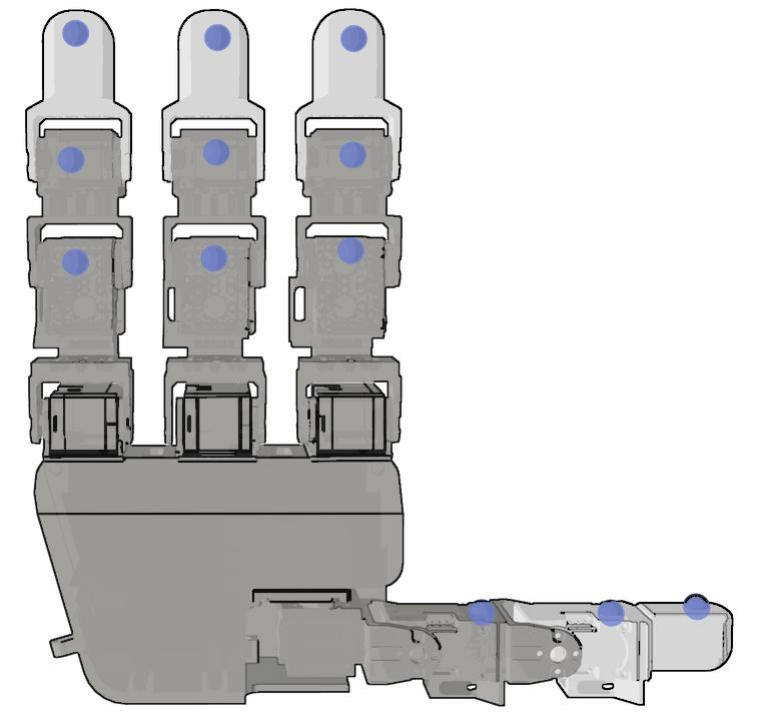}
    \vspace{-12pt} 
    \caption{\textbf{Hand Keypoint Representation.} Sparse surface keypoints on each finger.}
    \vspace{-5pt}
    \label{fig:keypoint_visualization}
\end{wrapfigure}

To represent contact interactions in a hand--object manipulation, we first construct a discrete representation of the object surface regions and the hand fingers. For the object, we uniformly sample $M$ surface points $\{(\mathbf{p}_m,\mathbf{n}_m)\}_{m=1}^M$, where each point encodes position and surface normal. We initialize $K$ region centers with farthest-point sampling and refine them with K-means using a geometry-aware distance that combines positional proximity and normal agreement:
\begin{equation}\label{eq:surface_region_distance}
    d_{\mathrm{geom}}(m,k)
    =(1-\lambda_{\mathrm{n}})\,\lVert\mathbf{p}_m-\boldsymbol{\mu}_k\rVert_2
    +\lambda_{\mathrm{n}}\bigl(1-\mathbf{n}_m^{\top}\bar{\mathbf{n}}_k\bigr),
\end{equation}
where $\boldsymbol{\mu}_k$ and $\bar{\mathbf{n}}_k$ are the position and mean normal of region $k$, and $\lambda_{\mathrm{n}}$ weights normal similarity. Each point is assigned to $\xi(m)=\arg\min_k d_{\mathrm{geom}}(m,k)$, yielding the $K$ surface regions used by the contact coverage counter.

For the hand, we represent the hand as $F$ fingers and let $\mathcal{L}_f$ denote the index set of predefined surface keypoints on finger $f$. Each finger is abstracted by keypoint positions and corresponding normals $\{(\mathbf{p}_{\ell},\mathbf{n}_{\ell})\}_{\ell\in\mathcal{L}_f}$ on its links. These keypoints are attached to each finger link and located on their surfaces (Fig.~\ref{fig:keypoint_visualization}). We place all the keypoints on the palmar surfaces by default for simplicity, though this is not required; the resulting contact representation is robust to moderate perturbations toward the lateral sides of the fingers.
This sparse representation follows prior work~\cite{drograsp} showing that a limited number of contact points suffices to capture hand--object interaction dynamics across diverse manipulation tasks.

\begin{figure*}[!t]
    \centering
    \includegraphics[width=\textwidth]{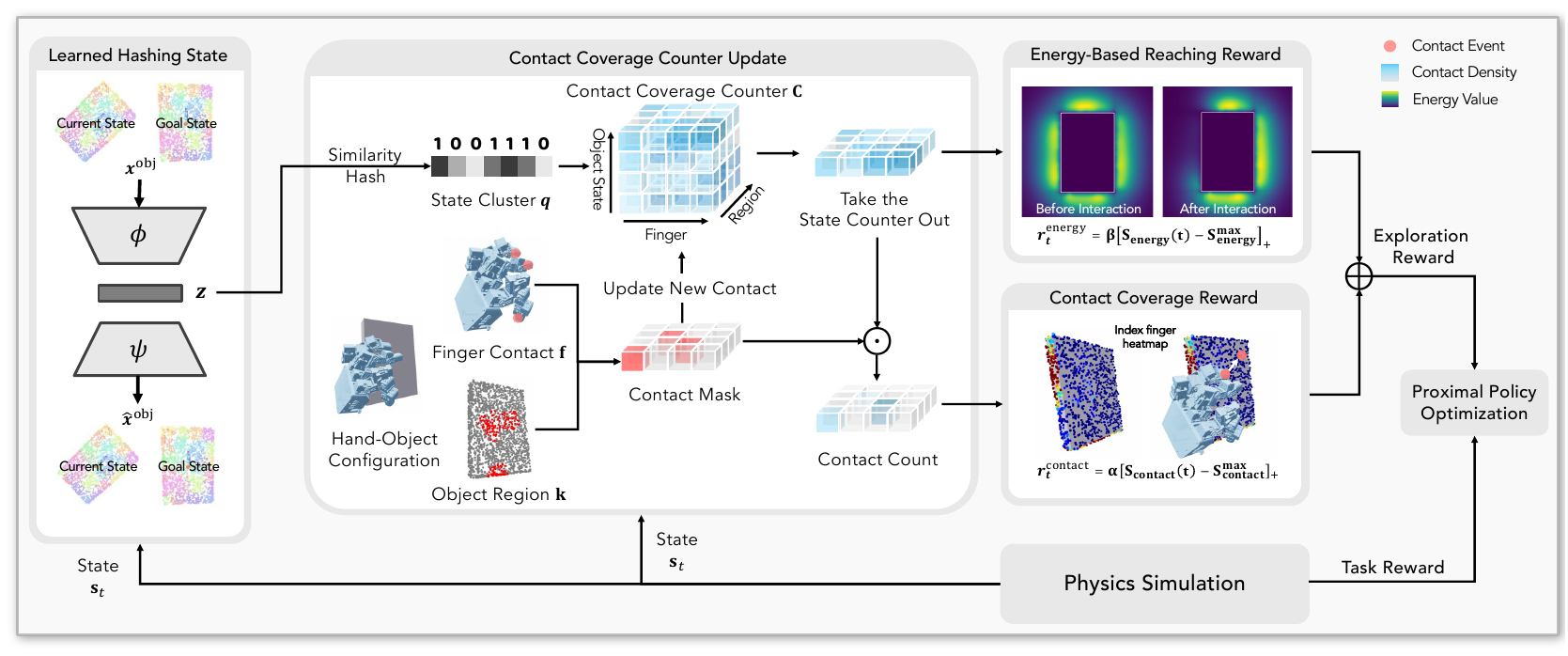}
    \caption{\textbf{Overview of ContactExplorer.} Our contact coverage-guided exploration framework consists of three main components: (1) a learned state hashing module that clusters continuous object states (Sec.~\ref{sec:state_cluster}); (2) a contact coverage counter that tracks state-conditioned finger-region interactions (Sec.~\ref{sec:ccc}); and (3) a structured exploration reward mechanism, decomposed into contact coverage (Sec.~\ref{sec:contct_coverage_reward}) and energy-based reaching terms (Sec.~\ref{sec:energe_reaching_reward}). In the visualization, the current and goal object states are depicted as distinctly colored point clouds.}
    \label{fig:ContactExplorer}
    \vspace{-4mm}
\end{figure*}

\subsection{Object State Cluster via Learned Hashing}
\label{sec:state_cluster}

In dexterous manipulation, an identical contact pattern may be useful across different spatial configurations and temporal phases of the object. Maintaining a single global visitation counter causes previously explored patterns to be treated as non-novel regardless of the underlying object state, thereby discouraging the reuse of potentially meaningful interactions under new spatiotemporal contexts. 

To enable state-aware exploration, we condition visitation statistics on the object state. At time step $t$, we define the object state as $\mathbf{x}^{\text{obj}}_t = [\mathbf{x}^{\text{cur}}_t, \mathbf{x}^{\text{goal}}_t]$, where both terms are point-cloud representations obtained by transforming a fixed set of $M$ canonical surface points according to the object's current and goal poses. We partition the continuous state space into $2^H$ discrete state clusters and denote the cluster index at time $t$ by $q_t$. An independent contact counter is pre-allocated for each cluster, ensuring decoupled exploration statistics across different object states.

To construct these clusters, we adopt a learned hashing mechanism following Tang et al.~\cite{NIPS2017hash}. The encoder $\phi$ maps the object state to a continuous latent code $\mathbf{z}_t=\phi(\mathbf{x}^{\text{obj}}_t)\in(0,1)^D$, and the decoder $\psi$ reconstructs $\hat{\mathbf{x}}^{\text{obj}}_t=\psi(\mathbf{z}_t)=\psi(\phi(\mathbf{x}^{\text{obj}}_t))$. The autoencoder is trained with a reconstruction loss and a binarization regularizer:
\begin{align}\label{eq:ae_loss}
\mathcal{L}_{\text{AE}}
= \big\| \hat{\mathbf{x}}^{\text{obj}}_t  - \mathbf{x}^{\text{obj}}_t \big\|_2^2
    + \frac{\lambda}{D} \sum_{i=1}^{D}
    \min\!\left\{
        (1 - z_{t,i})^2,\;
        z_{t,i}^2
    \right\},
\end{align}
where $\hat{\mathbf{x}}^{\text{obj}}_t$ is the reconstruction of $\mathbf{x}^{\text{obj}}_t$, and $z_{t,i}$ is the $i$-th dimension of the latent code.
We then binarize $\mathbf{z}_t$ via thresholding at $0.5$ and project it to a compact $H$-bit hash using SimHash~\cite{SimHash} with a fixed random projection matrix. The resulting hash is the discrete cluster index $q_t \in \{0, \ldots, 2^{H} - 1\}$. This enables tracking of contact coverage across semantically similar object states. The autoencoder is trained jointly with policy optimization. During interaction, each contact event updates the counter $\mathbf{C}_{q_t,f,k}$ corresponding to the inferred cluster $q_t$ (Section~\ref{sec:ccc}).

\subsection{Contact Coverage Counter}
\label{sec:ccc}

To encourage exploration of novel contact patterns, we maintain a pre-allocated contact coverage counter $\mathbf{C} \in \mathbb{N}_0^{2^H \times F \times K}$, where $\mathbf{C}_{q,f,k}$ records contacts between finger $f$ and surface region $k$ under cluster index $q$. Contact is detected at the finger level (any keypoint contact triggers the finger), and this merged design empirically outperforms per-keypoint counters. During training, contact detection operates at the level of hand keypoints and object surface points. At time step $t$, for each finger $f$, we identify the closest interacting pair:
\begin{equation}\label{eq:point_pairs}
    (\ell_{t,f},m_{t,f}) = \mathop{\arg\min}\limits_{\ell\in\mathcal{L}_f,\;m\in\{1,\ldots,M\}} \|\mathbf{p}_{t,\ell} - \mathbf{p}_{t,m}\|_2,
\end{equation}
where $\ell_{t,f}$ is the keypoint on finger $f$ nearest to the object and $m_{t,f}$ is its closest object surface point at time step $t$, which is mapped to the surface region $k_{t,f}=\xi(m_{t,f})$. For brevity, we write $\mathbf{p}^{\star}_{t,f}\triangleq\mathbf{p}_{t,\ell_{t,f}}$ for the position of this active keypoint. We then register a finger-level contact indicator
\begin{equation}\label{eq:contact_detect_func}
    c_{t,f} = \mathbf{1}\!\left\{\text{finger } f \text{ is in contact with region } k_{t,f}\right\} \in \{0,1\},
\end{equation}
which marks the finger--region pair $(f,k_{t,f})$ as active. In practice $c_{t,f}$ is obtained from a simple distance--force criterion on the nearest keypoint pair, with the concrete thresholds given in the experimental setup (Section~\ref{sec:results}). We refer to the procedure in Equations~\ref{eq:point_pairs}--\ref{eq:contact_detect_func}, which computes $c_{t,f}$, $k_{t,f}$, and $\mathbf{p}^{\star}_{t,f}$, as \textsc{ContactQuery} in Algorithm~\ref{alg:cceg_compact}.
After the exploration rewards at time step $t$ are evaluated using the current counts, $\mathbf{C}_{q_t,f,k_{t,f}}$ is incremented by one whenever $c_{t,f}=1$. This post-reward update ensures that a first visit is evaluated with count zero. The discrete approximation avoids expensive fine-grained collision detection while remaining robust to simulation noise. Counters are maintained without reset throughout RL training, enabling a natural transition from exploration to exploitation.

\section{Contact Coverage-Guided Exploration}
\label{sec:method}

\para{Design overview} Contact events are inherently sparse during manipulation: rewarding only novel contact patterns provides no guidance for motion in free space, while generic state-novelty objectives often encourage manipulation-irrelevant behaviors in free space. To address this, ContactExplorer decomposes exploration into two complementary signals derived from the same contact coverage statistics. The \textit{post-contact} signal provides a sparse, contact-focused reward that rewards only interaction-relevant behaviors. The \textit{pre-contact} signal provides dense, continuous guidance by shaping motion toward spatial regions likely to yield novel contact patterns. Together, they deliver structured exploration during the training process. Fig.~\ref{fig:ContactExplorer} illustrates the overall pipeline.

\subsection{Contact Coverage Reward}
\label{sec:contct_coverage_reward}
\para{Post-contact novelty} To ensure exploration remains focused on interaction-relevant events, we provide intrinsic rewards only upon physical contact. At time step $t$, for each finger $f$ that contacts the object ($c_{t,f}=1$), we map the contacted point to its corresponding surface region $k_{t,f}$ under the current cluster index $q_t$, and compute a contact novelty score:
\begin{align}\label{eq:rew_contact}
S_{\text{contact}}(t)
=
\frac{1}{F}
\sum_{f=1}^{F}
c_{t,f}
\cdot
w(\mathbf{C}_{q_t,f,k_{t,f}}),
\end{align}
where $w:\mathbb{N}_0\to(0,1]$, $w(n)=(n+1)^{-1/2}$, is a monotonically decreasing novelty weight for contact count $n$~\cite{NIPS2016unify}. Thus, first-time finger--region contacts receive the largest bonus, while repeated contacts contribute progressively less.
Both exploration scores at time step $t$ are evaluated before the current contacts are added to the counter.

To avoid repeatedly exploiting previously discovered interaction trajectories, we further adopt a progress-based shaping scheme that rewards only improvements over the best previously achieved score within the current episode:
\begin{align}
r_t^{\text{contact}}
=
\alpha
\left[
S_{\text{contact}}(t)
-
S_{\text{contact}}^{\max}
\right]_+,
\end{align}
where $S_{\text{contact}}^{\max}$ denotes the maximum contact novelty score previously achieved in the current episode, $\alpha$ is a scaling coefficient, and $[x]_+=\max(x,0)$ denotes the positive-part operator.

\subsection{Energy-Based Reaching Reward}
\label{sec:energe_reaching_reward}
\para{Pre-contact guidance} While the contact coverage reward encourages exploration after contact occurs, it provides no guidance before contact is established. To facilitate pre-contact exploration, we introduce an energy-based reaching objective that guides the policy toward under-explored contact patterns.

For each finger $f$, we define a contact energy:
\begin{align}
\Phi_{t,f}
=
\sum_{m=1}^{M}
\underbrace{w(\mathbf{C}_{q_t,f,\xi(m)})}_{\text{novelty weight}}
\,
\underbrace{\exp\!\left(
-
\frac{
\|
\mathbf{p}^{\star}_{t,f}
-
\mathbf{p}_{t,m}
\|_2
}{\sigma}
\right)}_{\text{proximity kernel}},
\end{align}
where $\sigma$ controls the spatial decay, $\mathbf{p}^{\star}_{t,f}$ denotes the active keypoint position on finger $f$, and $\mathbf{p}_{t,m}$ denotes an object-surface point at time step $t$. The resulting potential is larger near nearby regions with low contact counts, providing guidance before physical contact is established.

The energy-based exploration score is then defined as:
\begin{align}
S_{\text{energy}}(t)
=
\frac{1}{F}
\sum_{f=1}^{F}
\Phi_{t,f}.
\end{align}

Similar to the contact reward, we apply episodic progress-based shaping:
\begin{align}
  \label{eq:rew_energy}
r_t^{\text{energy}}
=
\beta
\left[
S_{\text{energy}}(t)
-
S_{\text{energy}}^{\max}
\right]_+,
\end{align}
where $S_{\text{energy}}^{\max}$ denotes the maximum energy score achieved within the current episode and $\beta$ is a scaling coefficient.

\para{Training procedure} The complete update is summarized in Algorithm~\ref{alg:cceg_compact}. The counter is persistent during training, while the episodic maxima reset at the beginning of each episode.

\begin{algorithm}[t]
\caption{Contact Coverage-Guided Exploration}
\label{alg:cceg_compact}
\begin{algorithmic}
\REQUIRE Surface points $\{\mathbf{p}_m\}_{m=1}^M$, encoder $\phi$, decoder $\psi$, policy $\pi$, value $V$; novelty weight $w(n)=(n+1)^{-1/2}$, $\sigma$, $\alpha,\beta$, task reward $r_t^{\text{task}}$.
\STATE Initialize $\mathbf{C}\in\mathbb{N}_0^{2^H\times F\times K}\gets\mathbf{0}$, persistent across episodes.
\WHILE{not converged}
  \STATE Reset $S_{\text{contact}}^{\max},S_{\text{energy}}^{\max}\gets0$ for the new episode
  \FOR{step $t=1,\ldots,T$}
    \STATE $q_t\gets\text{SimHash}(\mathbf{1}\{\phi(\mathbf{x}^{\text{obj}}_t)>0.5\})$
    \FOR{finger $f=1,\ldots,F$}
      \STATE $(c_{t,f},k_{t,f},\mathbf{p}^{\star}_{t,f})\gets\textsc{ContactQuery}(t,f)$
      \STATE $\Phi_{t,f}\gets\sum_{m=1}^{M}w(\mathbf{C}_{q_t,f,\xi(m)})\exp(-\|\mathbf{p}^{\star}_{t,f}-\mathbf{p}_{t,m}\|_2/\sigma)$
    \ENDFOR
    \STATE $S_{\text{contact}}(t)\gets \frac{1}{F}\sum_f c_{t,f}w(\mathbf{C}_{q_t,f,k_{t,f}})$
    \STATE $S_{\text{energy}}(t)\gets \frac{1}{F}\sum_f\Phi_{t,f}$
    \STATE $r_t^{\text{contact}}\gets\alpha[S_{\text{contact}}(t)-S_{\text{contact}}^{\max}]_+$; update $S_{\text{contact}}^{\max}$
    \STATE $r_t^{\text{energy}}\gets\beta[S_{\text{energy}}(t)-S_{\text{energy}}^{\max}]_+$; update $S_{\text{energy}}^{\max}$
    \STATE $r_t\gets r_t^{\text{task}}+r_t^{\text{contact}}+r_t^{\text{energy}}$
    \FOR{finger $f=1,\ldots,F$}
      \STATE $\mathbf{C}_{q_t,f,k_{t,f}}\gets\mathbf{C}_{q_t,f,k_{t,f}}+c_{t,f}$
    \ENDFOR
    \STATE Update $\pi,V$ with PPO; update $\phi,\psi$ using Eq.~\eqref{eq:ae_loss}
  \ENDFOR
\ENDWHILE
\end{algorithmic}
\end{algorithm}

\section{Experiments}
\label{sec:results}

\begin{figure}[!t]
    \centering
    \includegraphics[width=\linewidth]{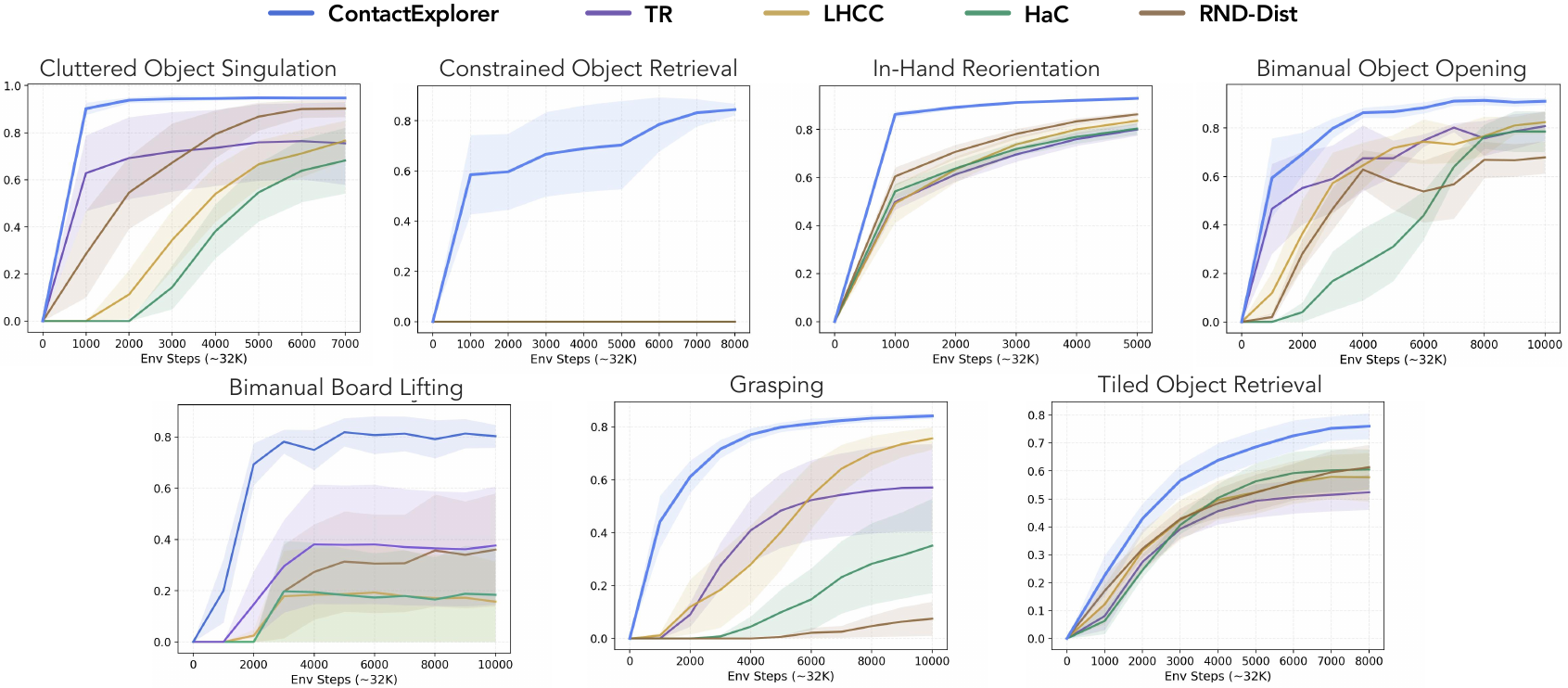}
    \caption{\textbf{Learning curves on seven dexterous manipulation tasks.} Each panel plots success rate against environment steps for ContactExplorer and four baselines (\method{TR}, \method{LHCC}, \method{HaC}, \method{RND-Dist}), averaged over five random seeds; shaded regions denote $\pm1$ standard deviation across seeds. ContactExplorer attains higher final success rates and faster convergence, and is the only method to make progress on the hard-exploration \task{Constrained Object Retrieval} task, where all baselines fail.}
    \label{fig:curve}
    \vspace{-4mm}
\end{figure}

\begin{figure}[!t]
    \centering
    \includegraphics[width=\linewidth]{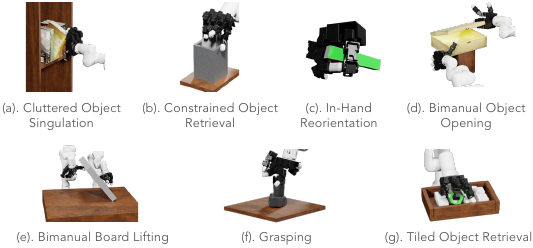}
    \caption{\textbf{Task overview.} Seven contact-rich tasks, each demanding a distinct interaction pattern. \textbf{(a)}~\task{Cluttered Object Singulation}: extract a target from a densely packed row. \textbf{(b)}~\task{Constrained Object Retrieval}: retrieve an object from a walled space. \textbf{(c)}~\task{In-Hand Reorientation}: rotate an object to a goal orientation. \textbf{(d)}~\task{Bimanual Object Opening}: two hands open an articulated object. \textbf{(e)}~\task{Bimanual Board Lifting}: two hands lift an oversized board. \textbf{(f)}~\task{Grasping}: grasp and lift a target. \textbf{(g)}~\task{Tiled Object Retrieval}: retrieve a cube from a tightly tiled bin.}
    \label{fig:tasks}
    \vspace{-4mm}
\end{figure}

We ask whether ContactExplorer improves task performance and sample efficiency over existing exploration rewards (\textbf{Q1}), reduces reliance on task-specific priors (\textbf{Q2}), how its key design components contribute to overall performance (\textbf{Q3}), and whether it remains effective across hand embodiments and transfers from simulation to the real world (\textbf{Q4}).

\subsection{Experimental Setup}
\para{Tasks and Evaluation Protocol} We evaluate the seven representative contact-rich tasks shown in Fig.~\ref{fig:tasks}. Unless otherwise stated, experiments use a UFactory xArm with a 16-DoF LEAP Hand~\cite{shaw2023leaphand} in Isaac Gym~\cite{makoviychuk2021isaac} with PPO. Multi-embodiment experiments additionally use mjlab~\cite{mjlab} (Section~\ref{sec:realworldexperiments}). Across all experimental settings, we train with 2048 parallel environments and five random seeds, and report success rate (SR) at the full budget and at 40\% of training. Unless otherwise noted, simulation results are reported as mean $\pm$ standard deviation over the five seeds, with the best result in bold.

\para{Controlled Comparison} Object meshes are represented by canonical surface point clouds partitioned into regions using farthest-point initialization and K-means with positional and normal similarity. A finger contact is registered when the nearest keypoint is within \SI{0.5}{cm} of the object and its force exceeds \SI{0.01}{N}. Within each embodiment, all policies use the same proprioceptive, tactile, object-state, and previous-action inputs, low-level position-controlled action space, policy architecture, and hyperparameters; only the exploration reward differs. Embodiment-specific action dimensions, kinematics, and simulator parameters are configured according to the corresponding hand and simulation backend.

\subsection{Q1: Overall Performance and Sample Efficiency}
To answer \textbf{Q1}, we compare ContactExplorer with four baselines under identical policy architectures, observations, and hyperparameters: \method{TR}, using task reward only; \method{LHCC}, a learned-hash-code count-based method~\cite{NIPS2017hash,pmlr-v270-zhang25a-wococo}; \method{HaC}, based on haptic prediction error~\cite{hapticscuriosity}; and \method{RND-Dist}, applying random network distillation to hand--object distance~\cite{pmlr-v229-schwarke23a-curiosity}. The best SR and the SR at 40\% of training are reported together in Table~\ref{tab:main_results_sr}. We further show the learning curves in Fig.~\ref{fig:curve} to compare training efficiency across methods.

ContactExplorer achieves the highest average success rate, with the lowest seed variance averaged across tasks, and is the only method to solve \task{Constrained Object Retrieval}. It also improves sample efficiency: at only 40\% of the training budget it already reaches an average success rate of $75\%$, whereas every baseline stays at $47\%$ or below under the same budget (Table~\ref{tab:main_results_sr}). Consistently, the learning curves in Fig.~\ref{fig:curve} show ContactExplorer crossing each task's success threshold earlier than the exploration baselines, which plateau lower or require substantially more interaction. The baseline ordering matches the two failure modes of Section~\ref{sec:related_work}: \method{HaC} is weakest overall, as force prediction is unreliable under sparse and discontinuous contacts, while \method{RND-Dist} stays at or below \method{TR}, as distance novelty rewards approach without requiring contact.


\begin{table*}[t]
\caption{\textbf{Main results across all tasks under full and 40\% training budgets.} Per-task SR in \%, mean\,$\pm$\,std over five seeds; best in bold. Avg. is the unweighted mean of the seven per-task means.}
\label{tab:main_results_sr}
\renewcommand{\arraystretch}{1.15}
\resizebox{\linewidth}{!}{%
\begin{tabular}{llcccccccc}
\toprule
\multirow{2}{*}{Budget} & \multirow{2}{*}{Method} & \multicolumn{7}{c}{Task} & \multirow{2}{*}{Avg.} \\
\cmidrule(lr){3-9}
 & & \task{Singulation} & \makecell{\task{Constrained}\\\task{Retrieval}} & \makecell{\task{In-Hand}\\\task{Reorientation}} & \makecell{\task{Bimanual}\\\task{Opening}} & \makecell{\task{Board}\\\task{Lifting}} & \task{Grasping} & \makecell{\task{Tiled}\\\task{Retrieval}} & \\
\midrule
\multirow{5}{*}{40\%} & \method{TR} & \ms{72}{34} & \ms{0}{0} & \ms{64}{16} & \ms{74}{25} & \ms{27}{37} & \ms{52}{25} & \ms{39}{7} & $47$ \\
 & \method{LHCC}~\cite{NIPS2017hash,pmlr-v270-zhang25a-wococo} & \ms{34}{26} & \ms{0}{0} & \ms{64}{14} & \ms{67}{25} & \ms{18}{37} & \ms{44}{25} & \ms{42}{14} & $38$ \\
 & \method{HaC}~\cite{hapticscuriosity} & \ms{14}{19} & \ms{0}{0} & \ms{64}{13} & \ms{51}{38} & \ms{38}{47} & \ms{29}{23} & \ms{41}{6} & $34$ \\
 & \method{RND-Dist}~\cite{pmlr-v229-schwarke23a-curiosity} & \ms{67}{33} & \ms{0}{0} & \ms{66}{13} & \ms{69}{26} & \ms{19}{39} & \ms{27}{31} & \ms{43}{7} & $42$ \\
 \ourrow & \textbf{Ours} & \msb{94}{3} & \msb{67}{34} & \msb{76}{14} & \msb{85}{9} & \msb{75}{16} & \msb{72}{8} & \msb{57}{11} & $\mathbf{75}$ \\
\midrule
\multirow{5}{*}{Full} & \method{TR} & \ms{77}{33} & \ms{0}{0} & \ms{78}{12} & \ms{92}{5} & \ms{36}{44} & \ms{68}{21} & \ms{53}{12} & $58$ \\
 & \method{LHCC} & \ms{77}{17} & \ms{0}{0} & \ms{78}{12} & \ms{90}{7} & \ms{16}{31} & \ms{72}{9} & \ms{59}{16} & $56$ \\
 & \method{HaC} & \ms{68}{28} & \ms{0}{0} & \ms{77}{12} & \ms{88}{11} & \ms{38}{46} & \ms{61}{20} & \ms{61}{15} & $56$ \\
 & \method{RND-Dist} & \ms{91}{7} & \ms{0}{0} & \ms{79}{11} & \ms{86}{14} & \ms{18}{37} & \ms{44}{31} & \ms{62}{16} & $54$ \\
 \ourrow & \textbf{Ours} & \msb{96}{1} & \msb{88}{6} & \msb{86}{12} & \msb{96}{2} & \msb{80}{9} & \msb{79}{5} & \msb{77}{9} & $\mathbf{86}$ \\
\bottomrule
\end{tabular}%
}
\vspace{-0.1in}
\end{table*}


\subsection{Q2: Reducing Task-Specific Priors}

\begin{figure}[!t]
    \centering
    \includegraphics[width=0.84\linewidth]{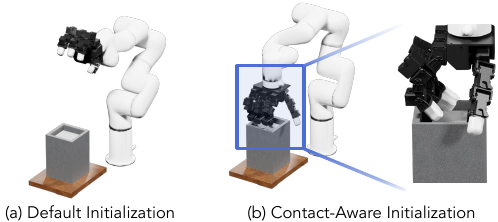}
    \caption{\textbf{Task-Specific Initialization Prior.} Default and contact-aware initialization in \task{Constrained Object Retrieval}.}
    \label{fig:task_prior}
\end{figure}

To answer \textbf{Q2}, we test whether ContactExplorer reduces the need for task-specific exploration priors on \task{Constrained Object Retrieval} by comparing it with \method{TR} and \method{TR-ContactInit}, which augments \method{TR} with a carefully designed contact-aware hand initialization that starts the hand near the object surface (Fig.~\ref{fig:task_prior}). As shown in Table~\ref{tab:q2_priors}, \method{TR} fails from the canonical initial state with 0\% SR, while \method{TR-ContactInit} partially succeeds with 33\% SR. ContactExplorer reaches 88\% SR without task-specific initialization and already achieves 67\% at 40\% of the training budget, compared with 25\% for \method{TR-ContactInit}. These results suggest that autonomously discovered contact patterns can replace manually designed priors while improving both performance and learning efficiency.


%

\subsection{Q3: Which Design Components Matter?}

To answer \textbf{Q3}, we first isolate the post-contact coverage and pre-contact energy rewards, and then evaluate whether state conditioning prevents contact-count saturation across distinct object configurations.

\para{Reward ablation} Table~\ref{tab:q3_reward_component_ablation} isolates the two exploration rewards on \task{Cluttered Object Singulation} and \task{Constrained Object Retrieval}. The complete method performs best overall, indicating that the post-contact coverage reward $r_t^{\text{contact}}$ and the pre-contact energy reward $r_t^{\text{energy}}$ are complementary.

\begin{table}[t]
    \centering
    \small
    \setlength{\tabcolsep}{6pt}
    \renewcommand{\arraystretch}{1.15}
    \caption{\textbf{Comparison with Task-Specific Prior.} SR in \%, mean\,$\pm$\,std over five seeds; best in bold.}
    \label{tab:q2_priors}
    \begin{tabularx}{\columnwidth}{lCC}
        \toprule
        Method & 40\% budget & Full budget \\
        \midrule
        \method{TR} & \ms{0}{0} & \ms{0}{0} \\
        \method{TR-ContactInit} & \ms{25}{32} & \ms{33}{40} \\
        \ourrow \textbf{Ours} & \msb{67}{34} & \msb{88}{6} \\
        \bottomrule
    \end{tabularx}
    \vspace{-0.08in}
\end{table}

\begin{table}[!t]
    \centering
    \small
    \setlength{\tabcolsep}{3pt}
    \renewcommand{\arraystretch}{1.15}
    \caption{\textbf{Reward Ablation Studies.} Sing. and Constrained Retri. denote Cluttered Object Singulation and Constrained Object Retrieval, respectively. SR in \%, mean\,$\pm$\,std over five seeds; best in bold.}
    \label{tab:q3_reward_component_ablation}
    \begin{tabular}{lcccc}
        \toprule
        \multirow{2}{*}{Reward terms} & \multicolumn{2}{c}{40\% budget} & \multicolumn{2}{c}{Full budget} \\
        \cmidrule(lr){2-3}\cmidrule(lr){4-5}
            & \task{Sing.} & \makecell{\task{Constrained}\\\task{Retri.}} & \task{Sing.} & \makecell{\task{Constrained}\\\task{Retri.}} \\
        \midrule
        Energy only & \ms{88}{4} & \ms{12}{15} & \ms{94}{2} & \ms{36}{44} \\
        Contact only & \ms{73}{23} & \ms{31}{38} & \ms{89}{7} & \ms{36}{44} \\
        \ourrow \textbf{Ours} & \msb{94}{3} & \msb{67}{34} & \msb{96}{1} & \msb{88}{6} \\
        \bottomrule
    \end{tabular}
    \vspace{-0.08in}
\end{table}

\begin{figure}[!t]
    \centering
    \includegraphics[width=\linewidth]{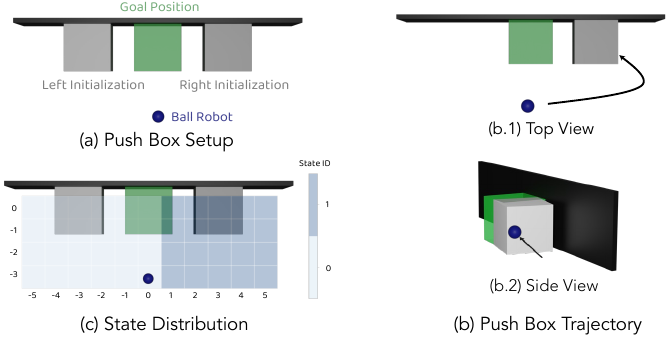}
    \caption{\textbf{\task{Box Push} diagnostic.} Learned state clusters keep the contact statistics of the two initial configurations separate.}
    \label{fig:push_box_main}
\end{figure}

\para{State-conditioned counter} We further evaluate object-state conditioning with \task{Box Push} (Fig.~\ref{fig:push_box_main}), a diagnostic task separate from the seven above, in which the box starts in either of two configurations requiring opposite pushing directions. Both configurations demand contacts on overlapping parts of the surface, so a single global counter treats those accumulated in one as already covering the other and stops rewarding them where they remain necessary. ContactExplorer achieves 100\% success, compared with 18\% without object-state conditioning, which typically solves only one configuration. The result also motivates state conditioning over heuristic counter decay, which requires hand-crafted schedules and may either discard useful counts or retain stale ones; ContactExplorer instead keeps separate counts for object states discovered online, using state transitions as an implicit signal of task progress.

\subsection{Q4: Generality Across Embodiments and Sim-to-Real Transfer}
\label{sec:realworldexperiments}
To answer \textbf{Q4}, we test whether the exploration signal remains effective when the hand embodiment changes, and whether the resulting policies transfer from simulation to a real robot.

\begin{figure}[!t]
    \centering
    \includegraphics[width=\columnwidth,trim=8 8 8 12,clip]{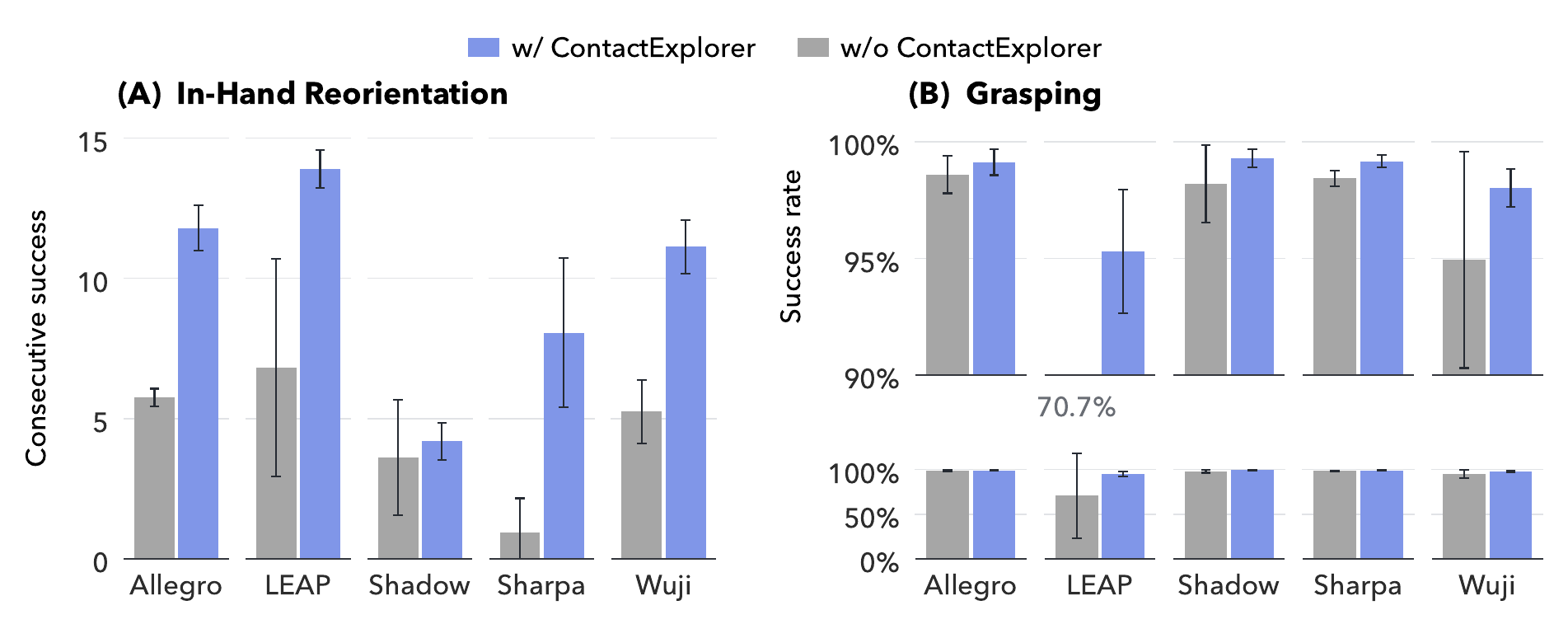}
    \caption{\textbf{Generality Across Embodiments.} Performance with (\emph{w/}) and without (\emph{w/o}) ContactExplorer across five dexterous hands on (A) \task{In-Hand Reorientation}, measured by consecutive successes, and (B) \task{Grasping}, measured by success rate. For grasping, the upper plot zooms in on the 90--100\% range, while the lower plot shows the full range. A separate policy is trained per hand. Bars show the mean over five seeds; error bars denote one standard deviation. ContactExplorer improves performance on every embodiment.}
    \label{fig:cross_embodiment}
\end{figure}

\para{Generality across embodiments} To examine whether the exploration signal is tied to a particular hand morphology, we train separate policies for five dexterous hands spanning different kinematics and finger counts---Allegro, LEAP, Shadow, Sharpa, and Wuji---on \task{In-Hand Reorientation} and \task{Grasping}, keeping the exploration method and training setup fixed while adapting only embodiment- and simulator-specific configurations. As shown in Fig.~\ref{fig:cross_embodiment}, ContactExplorer roughly doubles consecutive successes on \task{In-Hand Reorientation} for most hands. \task{Grasping} is near saturation without exploration, so margins are smaller there, with the clearest gain on LEAP (\SI{70.7}{\percent} to above \SI{95}{\percent}). These results suggest that the benefit of contact-coverage exploration is not specific to a single hand morphology.

%

\begin{figure}[!t]
    \centering
    \includegraphics[width=\linewidth]{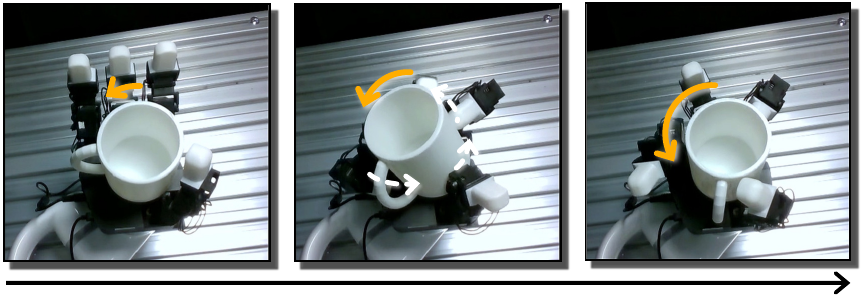}
    \caption{\textbf{Open-Loop Trajectory Replay.} Replaying privileged teacher actions open-loop on the real robot produces consistent \ang{90} $z$-axis rotations.}
    \label{fig:open-loop_trajectory_replay}
\end{figure}

\begin{figure}[!t]
    \centering
    \includegraphics[width=\linewidth]{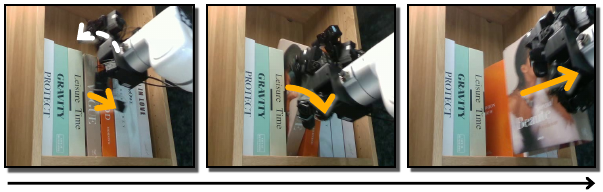}
    \caption{\textbf{Real-World Policy Execution.} A representative rollout (left to right) of the learned policy performing \task{Cluttered Object Singulation}.}
    \label{fig:real_world_deploy}
\end{figure}

\para{Sim-to-real transfer}
We deploy on a UFactory xArm with a 16-DoF LEAP Hand~\cite{shaw2023leaphand} and two RealSense D435 cameras, evaluating \task{In-Hand Reorientation} and \task{Cluttered Object Singulation}. A privileged teacher is trained with domain randomization, and 1,000 successful teacher trajectories are distilled into a visuomotor student using behavior cloning. The student receives masked 4D point clouds and proprioception, and predicts 6-DoF end-effector and 16-DoF hand commands at \SI{10}{Hz}. For \task{In-Hand Reorientation}, replaying privileged teacher actions produces consistent \ang{90} z-axis rotations on the real robot (Fig.~\ref{fig:open-loop_trajectory_replay}). We run 30 trials per policy and report success at two stages: \emph{exposure}, where at least half of the target object is exposed and grasp-feasible, and \emph{transport}, where the object is also moved to the goal. ContactExplorer improves both over \method{TR} on \task{Cluttered Object Singulation}; a representative execution sequence is shown in Fig.~\ref{fig:real_world_deploy}.

\begin{table}[t]
    \centering
    \small
    \setlength{\tabcolsep}{5pt}
    \caption{\textbf{Real-World \textnormal{\task{Cluttered Object Singulation}} Results.} Successes over 30 trials; \textnormal{\emph{Transport}} additionally requires moving the exposed object to the goal.}
    \label{tab:real_world_exp_results}
    \renewcommand{\arraystretch}{1.15}
    \begin{tabularx}{\columnwidth}{lCC}
        \toprule
        Method & Exposure & Transport \\
        \midrule
        \method{TR} & $11/30~(36.7\%)$ & $1/30~(3.3\%)$ \\
        \ourrow \textbf{Ours} & $\mathbf{23/30~(76.7\%)}$ & $\mathbf{10/30~(33.3\%)}$ \\
        \bottomrule
    \end{tabularx}
    \vspace{-0.2in}
\end{table}

Table~\ref{tab:real_world_exp_results} reports the real-world results. ContactExplorer raises exposure success from $36.7\%$ to $76.7\%$ and transport success from $3.3\%$ to $33.3\%$. The remaining gap between the two mainly reflects failures to catch and stabilize the object after it has been exposed.





\section{Limitations and Conclusion}
While ContactExplorer improves exploration across a broad range of dexterous manipulation tasks, several limitations remain. First, ContactExplorer is less suitable for tasks with cyclic object states, where a state recurs and the task requires repeating the same contact on each recurrence: the counter for that state is already incremented from earlier visits, so its novelty weight has decayed and the needed repetition earns little reward. Learned hashing helps only when the recurring states differ enough to fall into separate clusters. Second, the current exploration signal is mainly based on geometric contact coverage. Incorporating additional sensing modalities, such as force-torque or tactile sensing, could provide richer information about contact quality and interaction dynamics. Third, while ContactExplorer demonstrates promising sim-to-real transfer, most evaluations in this work are still conducted in simulation. Accelerating RL directly in the real world remains an important direction for future work.

In this work, we introduce \textbf{ContactExplorer}, a general exploration reward that explicitly incentivizes structured hand--object interactions for dexterous manipulation. By modeling contact coverage between fingers and object regions and combining sparse post-contact rewards with dense pre-contact guidance, it explores efficiently across diverse manipulation tasks without handcrafted shaping. Our experiments support contact coverage as a broadly applicable exploration component, and highlight the importance of state conditioning when contact patterns depend on object configuration.

\bibliographystyle{IEEEtran}
\bibliography{citation}

\clearpage
\begin{bibunit}[IEEEtran]
\makeatletter
\@startbibunitorrelax
\immediate\write\@bibunitaux{\string\citation{app:bst-control}}
\makeatother
\section*{Appendix}
\setcounter{section}{0}
\renewcommand{\thesection}{\Alph{section}}
\renewcommand{\thesectiondis}{\thesection.}
\renewcommand{\theHsection}{appendix.\Alph{section}}
\definecolor{appNovelty}{HTML}{235789}
\definecolor{appProximity}{HTML}{A65B16}
\definecolor{appDirection}{HTML}{70529A}
\definecolor{appVisibility}{HTML}{187565}
\newcommand{\appterm}[3]{\underbrace{\textcolor{#1}{#2}}_{\textcolor{#1}{\text{#3}}}}

\section{Implementation Details}
\label{app:implementation}

\para{Object and Hand Representations} Canonical object-surface points are partitioned by farthest-point initialization and geometry-aware clustering (Eq.~\eqref{eq:surface_region_distance}). They are transformed to the current object pose for contact queries and reaching-score evaluation, while their region assignments $\xi(m)$ index the counters. Hand keypoints are fixed to finger links, on palmar surfaces by default.

Surface points are indexed by $m$, keypoints on finger $f$ by $\ell\in\mathcal{L}_f$, and surface regions by $k=\xi(m)$. The object-state hash is $q_t$. The closest pair $(\ell_{t,f},m_{t,f})$ determines the active keypoint $\mathbf{p}^{\star}_{t,f}=\mathbf{p}_{t,\ell_{t,f}}$ and contacted region $k_{t,f}=\xi(m_{t,f})$. Current and goal point clouds are encoded and mapped through a fixed SimHash projection (Section~\ref{sec:state_cluster}).

\begin{figure*}[t]
\centering
\includegraphics[width=\textwidth]{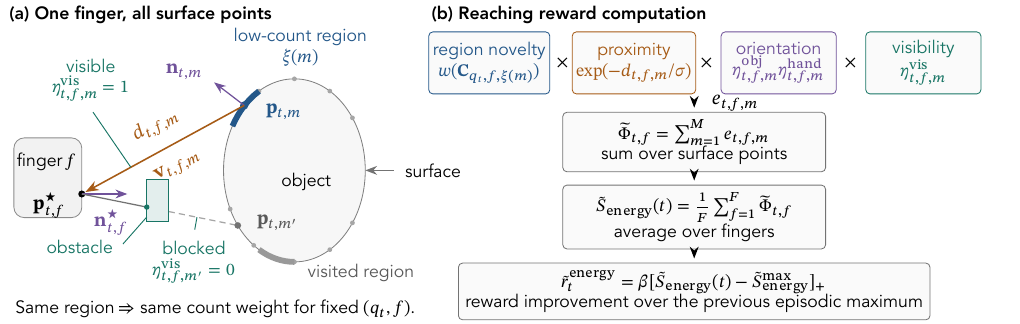}
\caption{\textbf{Reaching Energy Construction.} (a) Object surface normals and palmar-facing hand normals determine the directional weights. The approach to point $m$ is unblocked, while an obstacle masks point $m'$. (b) Point energies combine region novelty, proximity, orientation, and visibility, then are summed over surface points and averaged over fingers. The reaching reward measures improvement over the previous episodic maximum.}
\label{app:fig:energy-construction}
\end{figure*}

\para{Reaching Energy Construction} The reaching energy sums contributions from all $M$ surface points to the active keypoint of finger $f$. Each contribution uses the count for its surface region $\xi(m)$, weighted by proximity, orientation, and visibility (Fig.~\ref{app:fig:energy-construction}). The contact reward uses the count for the contacted region $k_{t,f}$.

Let $\mathbf{v}_{t,f,m}=\mathbf{p}^{\star}_{t,f}-\mathbf{p}_{t,m}$,
$d_{t,f,m}=\|\mathbf{v}_{t,f,m}\|_2$, and
$\mathbf{n}^{\star}_{t,f}=\mathbf{n}_{t,\ell_{t,f}}$.
All positions and normals in the following expressions are in the world frame.
Object normals $\mathbf{n}_{t,m}$ come from the canonical surface geometry
and rotate with the object. The hand normal $\mathbf{n}^{\star}_{t,f}$ is a
predefined palmar-facing direction in the keypoint's link frame, rotated
with that link.
Write $\widehat{\mathbf{v}}$ and $\widehat{\mathbf{n}}$ for normalized vectors.
The directional weights are positive parts of cosine similarities:
\begin{align}
\eta_{t,f,m}^{\mathrm{obj}}&=
[\widehat{\mathbf{v}}_{t,f,m}^{\top}\widehat{\mathbf{n}}_{t,m}]_+,\\
\eta_{t,f,m}^{\mathrm{hand}}&=
[-(\widehat{\mathbf{n}}^{\star}_{t,f})^{\top}\widehat{\mathbf{n}}_{t,m}]_+.
\end{align}
The first dot product compares the object normal with the direction toward
the finger: a surface facing away receives zero weight. The second compares
hand and object normals, giving its largest weight when they point in
opposite directions, as for opposing contact surfaces. Their product is
$\eta_{t,f,m}^{\mathrm{dir}}=\eta_{t,f,m}^{\mathrm{obj}}\eta_{t,f,m}^{\mathrm{hand}}\in[0,1]$.

Visibility tests the straight approach from $\mathbf{p}^{\star}_{t,f}$ to
$\mathbf{p}_{t,m}$. For cluttered singulation, obstacle boxes represent the
table and surrounding objects; for constrained retrieval, they represent
the container floor and walls. A ray--box intersection test checks whether
any box is encountered after leaving the keypoint and before reaching the
surface point. The binary visibility weight is
\begin{equation}
\eta_{t,f,m}^{\mathrm{vis}}=
\begin{cases}
0,&\text{an obstacle blocks the approach},\\
1,&\text{otherwise}.
\end{cases}
\label{app:eq:visibility-weight}
\end{equation}
A disabled geometric factor is set to one.
With novelty weight $w(n)=(n+1)^{-1/2}$, define
the point contribution
\begin{equation}
\begin{aligned}
e_{t,f,m}={}&\appterm{appNovelty}{w(\mathbf{C}_{q_t,f,\xi(m)})}{novelty}
\;\appterm{appProximity}{\exp(-d_{t,f,m}/\sigma)}{proximity}\\[4pt]
&\cdot\appterm{appDirection}{\eta_{t,f,m}^{\mathrm{dir}}}{orientation}
\;\appterm{appVisibility}{\eta_{t,f,m}^{\mathrm{vis}}}{visibility}.
\end{aligned}
\label{app:eq:point-energy}
\end{equation}
Thus, an occluded or back-facing point contributes zero even if it is close
to the finger and has a low contact count.
The geometrically weighted energy is
\begin{equation}
\widetilde{\Phi}_{t,f}=\sum_{m=1}^{M}e_{t,f,m}.
\label{app:eq:masked-energy}
\end{equation}
Setting $\eta_{t,f,m}^{\mathrm{dir}}=\eta_{t,f,m}^{\mathrm{vis}}=1$ recovers
$\Phi_{t,f}$. Averaging over fingers gives the energy score and its progress reward:
\begin{align}
\widetilde{S}_{\text{energy}}(t)&=\frac{1}{F}\sum_{f=1}^{F}\widetilde{\Phi}_{t,f},\\
\widetilde{r}_t^{\text{energy}}&=\beta[\widetilde{S}_{\text{energy}}(t)
-\widetilde{S}_{\text{energy}}^{\max}]_+.
\end{align}
Here $\widetilde{S}_{\text{energy}}^{\max}$ is the best score before step $t$
in the current episode. The reaching bonus rewards positive improvements over
this maximum.

\para{Training Settings} Tables~\ref{app:tab:ppo-settings} and~\ref{app:tab:contactexplorer-settings} list PPO hyperparameters and ContactExplorer parameters, respectively. We use five random seeds per setting. The object-state autoencoder is trained online with Eq.~\eqref{eq:ae_loss}.

\begin{table}[t]
\centering\rmfamily\normalsize
\setlength{\tabcolsep}{4pt}
\renewcommand{\arraystretch}{1.12}
\caption{\textbf{PPO Hyperparameters.}}
\label{app:tab:ppo-settings}
\begin{tabularx}{\columnwidth}{@{}lC@{}}
\toprule
Parameter & Value\\
\midrule
Parallel environments & 2,048\\
Learning-rate schedule & KL-adaptive\\
PPO epochs per update & 5\\
Minibatches per epoch & 4\\
Discount factor $\gamma$ & 0.99\\
GAE parameter $\lambda_{\mathrm{GAE}}$ & 0.95\\
Value-loss coefficient & 2.0\\
Entropy coefficient & 0\\
Maximum gradient norm & 1.0\\
Initial action standard deviation & 1.0\\
Activation & ELU\\
Observation normalization & Enabled\\
Advantage normalization & Across the rollout\\
\bottomrule
\end{tabularx}
\end{table}

\begin{table*}[t]
\centering\rmfamily\normalsize
\setlength{\tabcolsep}{5pt}
\renewcommand{\arraystretch}{1.12}
\caption{\textbf{ContactExplorer Parameters.} Manipulation includes single-arm and bimanual tasks. AE denotes the object-state autoencoder. $^{*}$Board lifting uses 1,024 surface points; object opening uses up to 1,024 per articulated part.}
\label{app:tab:contactexplorer-settings}
\begin{tabularx}{\textwidth}{lCCCC}
\toprule
\multirow{2}{*}{Parameter} & \multicolumn{2}{c}{Isaac Gym} & \multicolumn{2}{c}{mjlab}\\
\cmidrule(lr){2-3}\cmidrule(lr){4-5}
 & Manipulation & Reorientation & Reorientation & Grasping\\
\midrule
Contact coefficient $\alpha$ & 200 & 240 & 128 & 128\\
Reaching coefficient $\beta$ & 1.28 & 2.56 & 0.5 & 0.5\\
Hash length $H$ & 5 & 8 & 5 & 5\\
Surface points $M$ & 512\textsuperscript{*} & 512 & 512 & 512\\
Surface regions $K$ & 32 & 32 & 64 & 64\\
Latent dimension $D$ & 16 & 16 & 256 & 256\\
Reaching kernel scale $\sigma$ (m) & 0.03 & 0.03 & 0.03 & 0.03\\
\midrule
Binarization weight $\lambda$ & \multicolumn{4}{c}{1.0}\\
AE learning rate & \multicolumn{4}{c}{$3\times10^{-4}$}\\
AE hidden width & \multicolumn{4}{c}{512}\\
Latent noise amplitude & \multicolumn{4}{c}{0.3}\\
AE update interval (env. steps) & \multicolumn{4}{c}{16}\\
AE passes per update & \multicolumn{4}{c}{5}\\
AE minibatches per pass & \multicolumn{4}{c}{8}\\
\bottomrule
\end{tabularx}
\end{table*}

\section{Simulation Experimental Setup}
\label{app:simulation}

\para{Cluttered Object Singulation} Five equal-sized upright objects form a shelf row whose position is randomized along the shelf edge at scene creation. Only a randomly selected target is movable; the robot must extract it and move it to the goal.

\para{Constrained Object Retrieval} A cube sits below the rim of a top-opening container, with gaps too narrow for finger insertion. The robot must use interior-surface contacts to retrieve it and reach the goal.

\para{In-Hand Reorientation} Both up-facing and down-facing settings use sampled goal orientations and six ContactDB objects~\cite{Brahmbhatt_2019_CVPR_contactdb}: elephant, mug, bunny, duck, mouse, and teapot. Up-facing initial poses are randomized. Down-facing experiments additionally use a $16\times3\times3$\,cm cuboid; the six ContactDB objects are rescaled and initialized in fixed, grasped object--hand configurations.

\para{Bimanual Object Opening} Two hands support and open an articulated waffle iron or ARCTIC box~\cite{fan2023arctic}, moving its joint to the target angle.

\para{Bimanual Board Lifting} Two hands support opposite sides of a board too large for stable single-hand lifting and raise it to the target height.

\para{Tiled Object Retrieval} Eight cubes are tightly packed in a bin with walls higher than the cubes and gaps narrower than a finger. The robot must extract a target selected each episode and lift it $0.3$\,m above the table.

\para{Grasping} Rescaled camera, mug, and drill objects from OakInk~\cite{YangCVPR2022OakInk} start at random poses and must be grasped and transported to the goal.

Simulation reward terms and tolerances are given below; real-world exposure and transport criteria are defined separately in Section~\ref{app:real-world}.

\medskip
\noindent\textbf{Observation and Action Spaces}\par
\smallskip
For the main Isaac Gym experiments, let $\mathbf{T}_t^{b\mid c}$ denote the pose of body $b$ in frame $c$, with $\mathrm{w}$ denoting the world frame. Joint positions and velocities are $\mathbf{q}_t$ and $\dot{\mathbf{q}}_t$; superscripts identify the robot or hand. Policy observations concatenate the following features and the previous action $\mathbf{a}_{t-1}$.

\para{Single-arm manipulation} Inputs include robot joint states $(\mathbf{q}_t^{\mathrm{rob}},\dot{\mathbf{q}}_t^{\mathrm{rob}})$, end-effector pose and velocity, fingertip poses in the palm frame and their velocities, fingertip--object and object--palm relative poses, and target pose and velocity. Scene features comprise up to six nearest non-target object poses and the target bounding box. For each fingertip $f$, $d_{t,f}^{\mathrm{tip}}$ and $\widehat{\mathbf{u}}_{t,f}^{\mathrm{tip}}$ give the distance and unit direction to its nearest target-surface point. Inputs also include binary tactile signals $\mathbf{b}_t$. End-effector and object orientations use six-dimensional rotation features. The action $\mathbf{a}_t\in[-1,1]^{22}$ contains three translation, three rotation, and sixteen hand-joint velocity commands. Scaled Cartesian increments are converted to arm-joint targets by inverse kinematics; hand commands are integrated into position targets and clipped to joint limits. Singulation, constrained retrieval, tiled retrieval, and grasping use this feature set.

\para{In-hand reorientation} Inputs comprise hand joint positions $\mathbf{q}_t^{\mathrm{hand}}$, velocities $\dot{\mathbf{q}}_t^{\mathrm{hand}}$, and joint forces $\boldsymbol{\tau}_t^{\mathrm{hand}}$; object pose $\mathbf{T}_t^{\mathrm{obj}\mid\mathrm{w}}$ and velocity; goal pose $\mathbf{T}^{\mathrm{goal}\mid\mathrm{w}}$; and the quaternion error $\mathbf{r}_t^{\mathrm{err}}=\mathbf{r}_t^{\mathrm{obj}}\otimes(\mathbf{r}^{\mathrm{goal}})^{-1}$. Here $\mathbf{r}$ denotes a four-dimensional unit quaternion and $\otimes$ quaternion multiplication. For $n_h$ actuated hand joints, actions $\mathbf{a}_t\in[-1,1]^{n_h}$ are mapped linearly to joint-limit ranges as absolute position targets.

\para{Bimanual manipulation} Inputs include both hand-root poses and velocities, hand joint states, fingertip poses and velocities, object pose and velocity, object poses relative to each palm, and the left-hand pose relative to the right hand. Inputs also include binary tactile signals $\mathbf{b}_t$. Object opening additionally includes top- and bottom-part poses and velocities, articulated joint position $\theta_t^{\mathrm{arti}}$, and velocity $\dot{\theta}_t^{\mathrm{arti}}$. Orientations use six-dimensional rotation features. Actions $\mathbf{a}_t\in[-1,1]^{44}$ provide three translation, three rotation, and sixteen hand-joint velocity commands per hand; inverse kinematics and joint-velocity integration produce position targets for both arms and hands.

\medskip
\noindent\textbf{Task Rewards}\par
\smallskip
All compared methods share $r_t^{\mathrm{task}}$; ContactExplorer adds contact and reaching bonuses. Let $\mathbf{o}_t$ and $\mathbf{o}^{\mathrm{goal}}$ denote the object reference position and its target. The binary gate $g_t$ indicates whether the task's contact conditions are satisfied.

\para{In-hand reorientation} The task reward combines position and orientation tracking, an action penalty, and a success bonus:
\begin{equation}
\begin{split}
r_t^{\mathrm{task}}={}&w_p\|\mathbf{o}_t-\mathbf{o}^{\mathrm{goal}}\|_2
 +\frac{w_q}{|\theta_t|+\varepsilon}\\
 &+w_a\|\mathbf{a}_t\|_2^2+w_s\mathbf{1}\{|\theta_t|\leq\tau\}.
\end{split}
\label{app:eq:reorientation-reward}
\end{equation}
Here $\theta_t$ is the quaternion geodesic error, $\tau$ the success tolerance, $\varepsilon$ a stabilizer, and $\mathbf{a}_t$ the policy action. We use $w_p=-10$, $w_q=1$, $w_a=-2\times10^{-4}$, and $w_s=250$.

\para{Single-arm retrieval and grasping} Let $\mathcal{L}=\bigcup_{f=1}^{F}\mathcal{L}_f$ contain all hand-keypoint indices and $\mathcal{P}_t=\{\mathbf{p}_{t,m}\}_{m=1}^{M}$ be the target surface cloud. For keypoint position $\mathbf{p}_{t,\ell}$, define
\begin{align}
d_{t,\ell}&=\min_{\mathbf{p}\in\mathcal{P}_t}\|\mathbf{p}_{t,\ell}-\mathbf{p}\|_2,\\
d_t^{\mathrm{goal}}&=\|\mathbf{o}_t-\mathbf{o}^{\mathrm{goal}}\|_2.
\end{align}
Within each episode, $\bar d_{t,\ell}$ tracks the minimum keypoint distance, and $\bar d_t^{\mathrm{goal}}$ tracks the minimum goal distance when the contact gate is satisfied. The reach-progress term and task reward are
\begin{align}
\Delta_t&=\frac{1}{|\mathcal{L}|}\sum_{\ell\in\mathcal{L}}[\bar d_{t-1,\ell}-d_{t,\ell}]_+,\label{app:eq:reach-progress}\\
r_t^{\mathrm{task}}&=w_r\Delta_t+w_t g_t
 [\bar d_{t-1}^{\mathrm{goal}}-d_t^{\mathrm{goal}}]_+
 +r_t^{\mathrm{succ}}.\label{app:eq:retrieval-reward}
\end{align}
Keypoint and goal-distance minima start at $0.30$\,m and $0.60$\,m, respectively; $w_r=20$ and $w_t=60$. The term $r_t^{\mathrm{succ}}$ combines maintenance and success bonuses, with a success-bonus magnitude of 4,000. The goal indicator requires the contact gate and a position error within $0.075$\,m.

Tiled retrieval and grasping use $g_t=\mathrm{tip}_t\lor\mathrm{table}_t$; singulation uses $g_t=(\mathrm{tip}_t\lor\mathrm{table}_t)\land\neg\mathrm{env}_t$. Constrained retrieval uses
\begin{equation}
g_t=(\mathrm{tip}_t\lor\mathrm{floor}_t\lor\mathrm{wall}^{+y}_t)
\land\neg\mathrm{wall}^{\mathrm{other}}_t,
\label{app:eq:retrieval-contact-gate}
\end{equation}
Here $\mathrm{tip}_t$ denotes fingertip-keypoint contact with the target, $\mathrm{table}_t$ object--table contact, and $\mathrm{env}_t$ clutter contact. $\mathrm{floor}_t$, $\mathrm{wall}^{+y}_t$, and $\mathrm{wall}^{\mathrm{other}}_t$ denote target contact with the container floor, positive-$y$ wall, and remaining three walls, respectively. All are binary indicators.

\para{Bimanual manipulation} Reach progress is computed per hand against the relevant object part. With task progress $u_t$ and its contact-gated running maximum $\bar u_t$, the reward is
\begin{equation}
\begin{split}
r_t^{\mathrm{task}}={}&\frac{w_r^{\mathrm{bi}}}{2}
 (\Delta_t^{\mathrm{right}}+\Delta_t^{\mathrm{left}})\\
 &+w_t^{\mathrm{bi}}g_t[u_t-\bar u_{t-1}]_+
 +r_t^{\mathrm{succ}}.
\end{split}
\label{app:eq:bimanual-reward}
\end{equation}
For board lifting, $u_t$ is the lowest world-frame height of the board surface cloud; success requires a height within $0.05$\,m of the target and contacts from both hands. For object opening, $u_t=\theta_t^{\mathrm{arti}}/\theta^{\mathrm{goal}}$, with target angle $\theta^{\mathrm{goal}}=1.57$\,rad and tolerance $0.10$\,rad. The opening gate uses table contact and requires object-orientation displacement below $0.20$\,rad. Both tasks use $w_r^{\mathrm{bi}}=5$. The progress and success-bonus coefficients are $(300,4000)$ for lifting and $(100,200)$ for opening.

\para{Baselines and Evaluation Protocol} Baselines are task reward only (\method{TR}), learned-hash counts (\method{LHCC}), haptic prediction error (\method{HaC}), and random network distillation on hand--object distance (\method{RND-Dist}). Within each task and embodiment, methods share the policy architecture, observation/action spaces, task reward, and training budget.

We report simulation success rates as mean $\pm$ standard deviation over five seeds at 40\% and full training budgets. The aggregate success rate is the unweighted average of the seven task means. All sensitivity studies use Constrained Object Retrieval at these two checkpoints.


\section{Ablation and Sensitivity Studies}
\label{app:ablations}

\para{Exploration Reward Scales} We vary $\alpha$ and $\beta$ at a fixed ratio (Table~\ref{app:tab:reward-scales}). The smallest and largest tested scales yield lower final mean success rates than the default; doubling the default gives a similar mean (89\% versus 88\%).

\begin{table}[!t]
\centering\small
\setlength{\tabcolsep}{6pt}
\renewcommand{\arraystretch}{1.35}
\caption{\textbf{Exploration Reward Scales.} Constrained Object Retrieval SR in \%, mean\,$\pm$\,std over five seeds. Best mean per column in bold; the default is shaded.}
\label{app:tab:reward-scales}
\begin{tabularx}{\columnwidth}{>{\centering\arraybackslash}p{2.1em}>{\centering\arraybackslash}p{2.1em}CC}
\toprule
\multirow{2}{*}{$\alpha$} & \multirow{2}{*}{$\beta$} & \multicolumn{2}{c}{Success rate (\%)}\\
\cmidrule(lr){3-4}
 & & 40\% budget & Full budget\\
\midrule
$50$ & $0.32$ & \ms{14}{28} & \ms{17}{35}\\
$100$ & $0.64$ & \ms{30}{37} & \ms{52}{42}\\
\ourrow $200$ & $1.28$ & \ms{67}{34} & \ms{88}{6}\\
$400$ & $2.56$ & \msb{69}{16} & \msb{89}{4}\\
$800$ & $5.12$ & \ms{40}{35} & \ms{56}{45}\\
\bottomrule
\end{tabularx}
\end{table}

\para{Object-State Clustering} Hash length $H$ sets the maximum number of counter bins to $2^H$. Performance depends on both $H$ and the binarization weight $\lambda$ (Table~\ref{app:tab:state-clustering}): at $H=4$, raising $\lambda$ from 0.5 to 2 reduces the final mean success rate from 90\% to 52\%; at $H=6$, $\lambda=2$ yields the highest final mean in the sweep.

\begin{table}[!t]
\centering\small
\setlength{\tabcolsep}{6pt}
\renewcommand{\arraystretch}{1.35}
\caption{\textbf{Object-State Clustering Parameters.} Constrained Object Retrieval SR in \%, mean\,$\pm$\,std over five seeds. Best mean per column in bold; the default is shaded.}
\label{app:tab:state-clustering}
\begin{tabularx}{\columnwidth}{>{\centering\arraybackslash}p{2.1em}>{\centering\arraybackslash}p{2.1em}CC}
\toprule
\multirow{2}{*}{$H$} & \multirow{2}{*}{$\lambda$} & \multicolumn{2}{c}{Success rate (\%)}\\
\cmidrule(lr){3-4}
 & & 40\% budget & Full budget\\
\midrule
 & $0.5$ & \ms{68}{25} & \ms{90}{2}\\
 & $1.0$ & \ms{63}{32} & \ms{85}{6}\\
\multirow{-3}{*}{$4$} & $2.0$ & \ms{49}{40} & \ms{52}{43}\\
\cmidrule(lr){1-4}
 & $0.5$ & \ms{64}{33} & \ms{73}{36}\\
\ourrow & $1.0$ & \ms{67}{34} & \ms{88}{6}\\
\multirow{-3}{*}{$5$} & $2.0$ & \ms{62}{32} & \ms{72}{36}\\
\cmidrule(lr){1-4}
 & $0.5$ & \ms{57}{32} & \ms{71}{36}\\
 & $1.0$ & \ms{46}{37} & \ms{69}{35}\\
\multirow{-3}{*}{$6$} & $2.0$ & \msb{70}{10} & \msb{91}{4}\\
\bottomrule
\end{tabularx}
\end{table}

%
%
%
\section{Real-World Experimental Details}
\label{app:real-world}

\para{Hardware and Perception Setup} We use a UFactory xArm, a 16-DoF LEAP Hand, and two RealSense D435 RGB-D cameras. Depth pixels are back-projected with camera intrinsics and transformed to the robot base frame with calibrated extrinsics. The fused cloud is workspace-cropped, filtered for invalid depth, and reduced to 1,024 points by farthest-point sampling without replacement.

\begin{figure}[!t]
\centering
\begin{minipage}{0.48\linewidth}
\centering
\includegraphics[width=\linewidth]{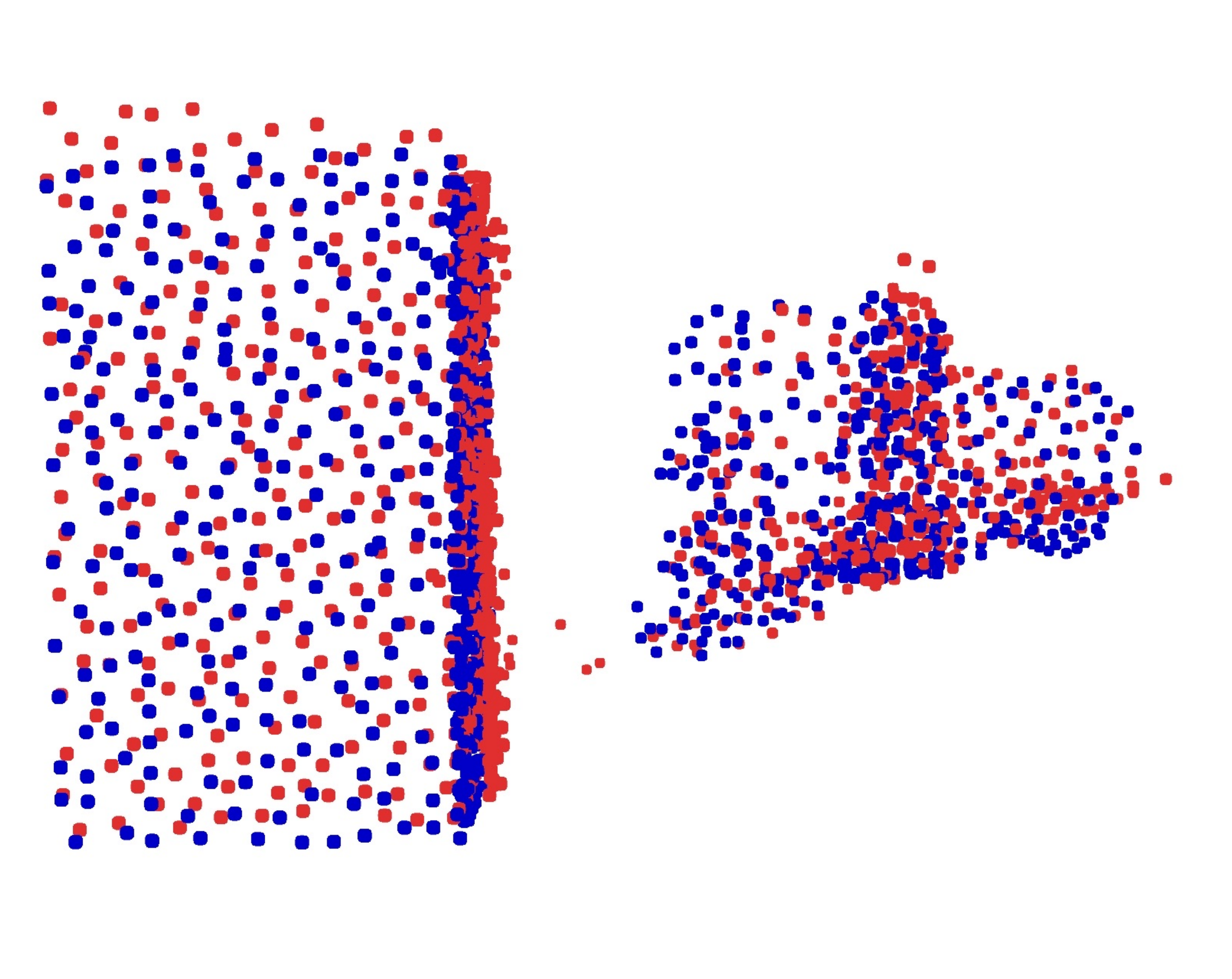}
{\footnotesize (a) Top view}
\end{minipage}\hfill
\begin{minipage}{0.48\linewidth}
\centering
\includegraphics[width=\linewidth]{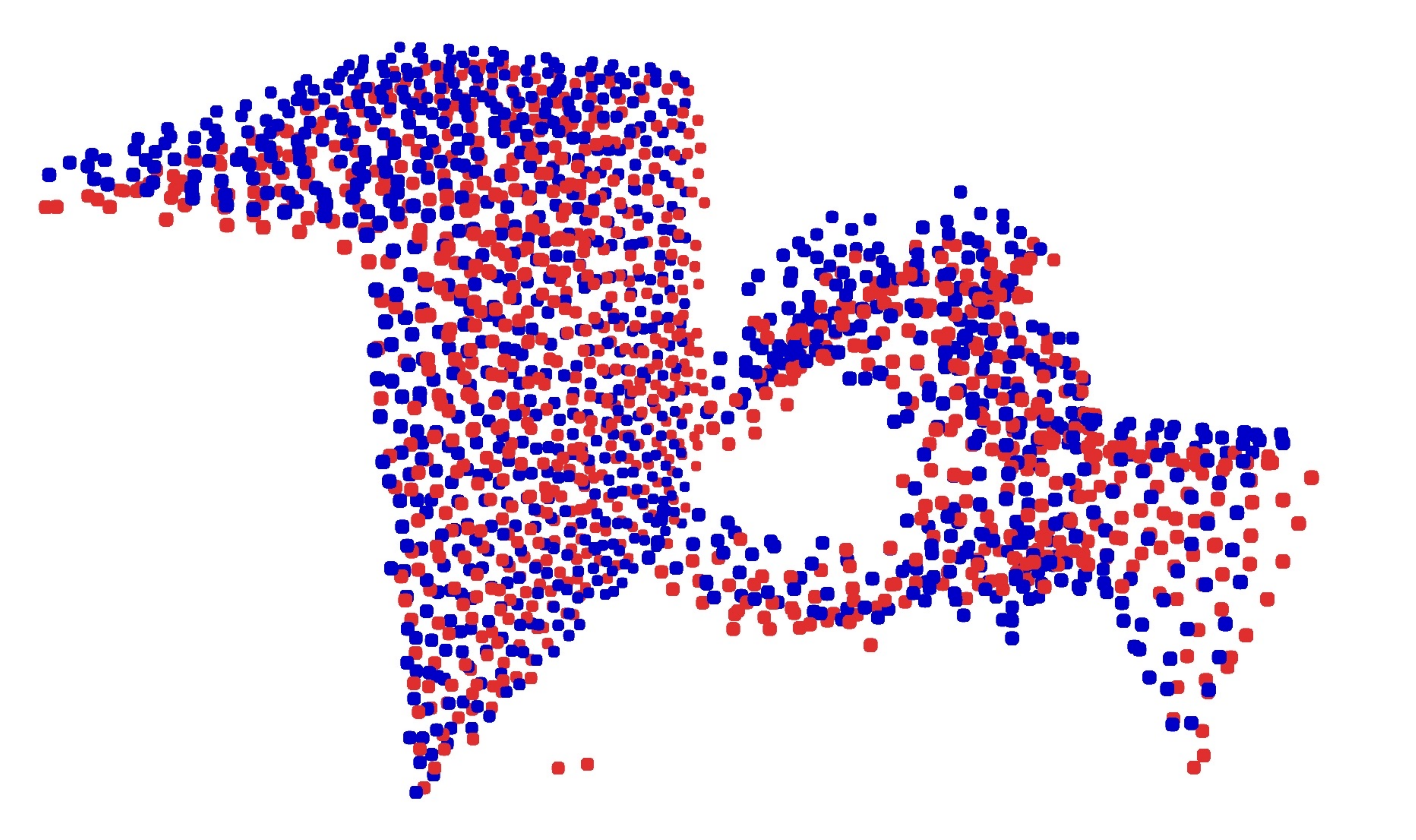}
{\footnotesize (b) Side view}
\end{minipage}
\caption{\textbf{Simulation and Real Point Clouds.} Simulated points are blue and real observations are red.}
\label{app:fig:pointclouds}
\end{figure}

SAM2~\cite{ravi2024sam2} supplies per-view target masks. The augmented scene cloud is $\mathbf{P}_t^{\mathrm{mask}}\in\mathbb{R}^{1024\times4}$, whose $i$-th row is $(x_{t,i},y_{t,i},z_{t,i},b_{t,i}^{\mathrm{mask}})$. Coordinates are expressed in the robot base frame; $b_{t,i}^{\mathrm{mask}}=1$ for target points and $0$ otherwise. Figure~\ref{app:fig:pointclouds} compares simulated and real clouds from two views.

\para{Teacher--Student Distillation} For visuomotor singulation, a privileged state-based teacher trained with domain randomization provides 1,000 successful trajectories with randomly sampled targets. The student learns by behavior cloning with L1 action loss.

PointNet~\cite{Qi2016PointNet} encodes $\mathbf{P}_t^{\mathrm{mask}}$; its features are concatenated with arm joint positions $\mathbf{q}_t^{\mathrm{arm}}\in\mathbb{R}^{7}$ and hand joint positions $\mathbf{q}_t^{\mathrm{hand}}\in\mathbb{R}^{16}$. The student uses a two-step history of each input (Table~\ref{app:tab:student-observations}) and predicts relative commands for the 6-DoF end-effector and 16 hand joints.

\begin{table}[!t]
\centering\rmfamily\normalsize
\setlength{\tabcolsep}{4pt}
\renewcommand{\arraystretch}{1.35}
\caption{\textbf{Student Observations.} Two-step histories of the scene point cloud with a binary target-mask channel and robot joint positions.}
\label{app:tab:student-observations}
\begin{tabularx}{\columnwidth}{@{}CC@{}}
\toprule
Observation & Dimension\\
\midrule
$\mathbf{P}_{t-1:t}^{\mathrm{mask}}$ & $2\times1024\times4$\\
$\mathbf{q}_{t-1:t}^{\mathrm{arm}}$ & $2\times7$\\
$\mathbf{q}_{t-1:t}^{\mathrm{hand}}$ & $2\times16$\\
\bottomrule
\end{tabularx}
\end{table}

\para{Deployment and Evaluation Protocol} The visuomotor policy runs at 10\,Hz with asynchronous sensing. Each step predicts an action from the latest available observations.

Singulation is evaluated over 30 trials per policy (Table~\ref{tab:real_world_exp_results}). \emph{Exposure} requires at least half the target to be exposed and grasp-feasible; \emph{transport} additionally requires reaching the goal. In-Hand Reorientation uses open-loop teacher-action replay for $90^{\circ}$ $z$-axis rotations (Fig.~\ref{fig:open-loop_trajectory_replay}).

\begin{figure}[!t]
\centering
\begin{minipage}{0.48\linewidth}
\centering
\includegraphics[width=0.90\linewidth]{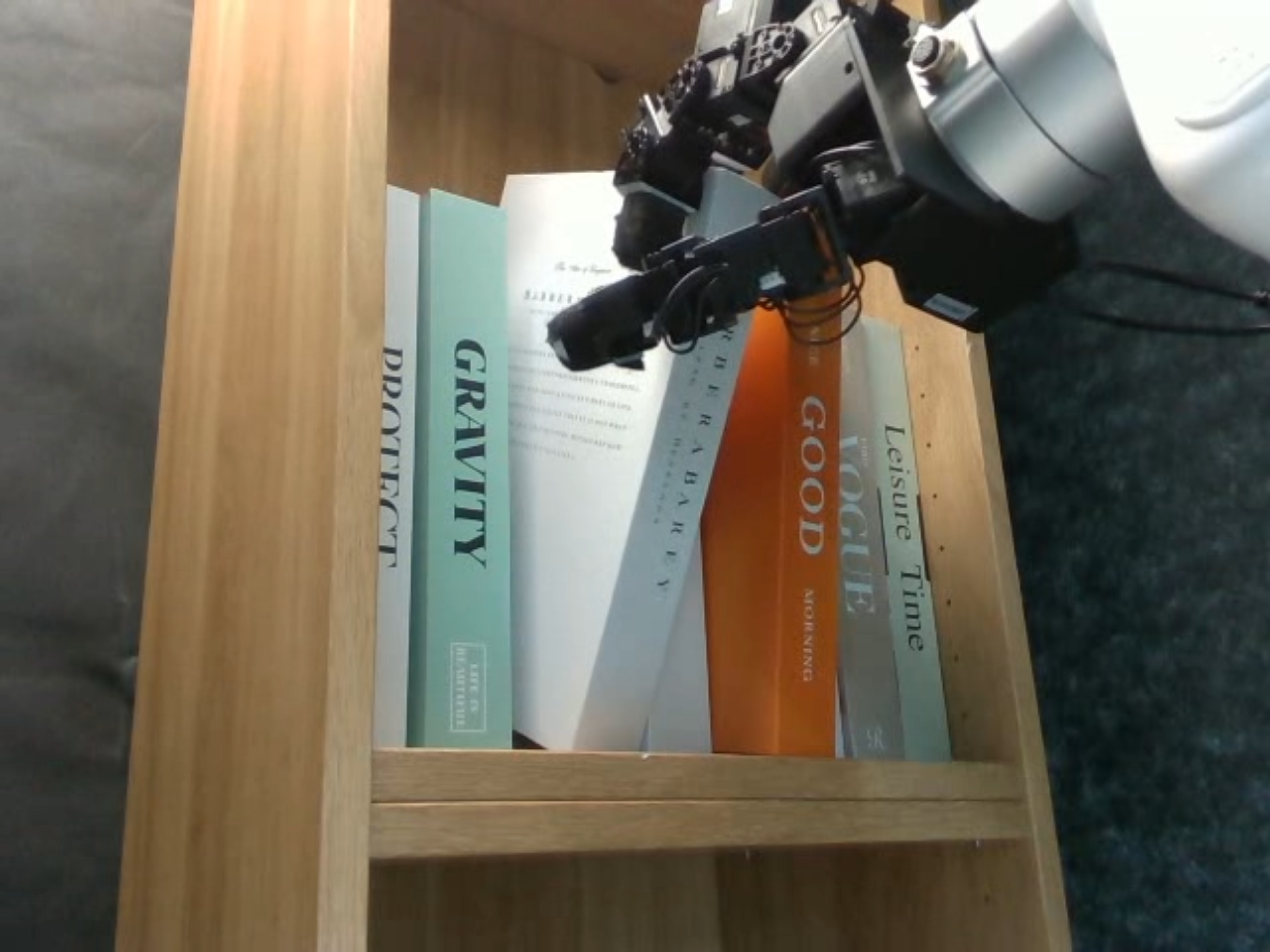}
{\footnotesize (a) Wrong target}
\end{minipage}\hfill
\begin{minipage}{0.48\linewidth}
\centering
\includegraphics[width=0.90\linewidth]{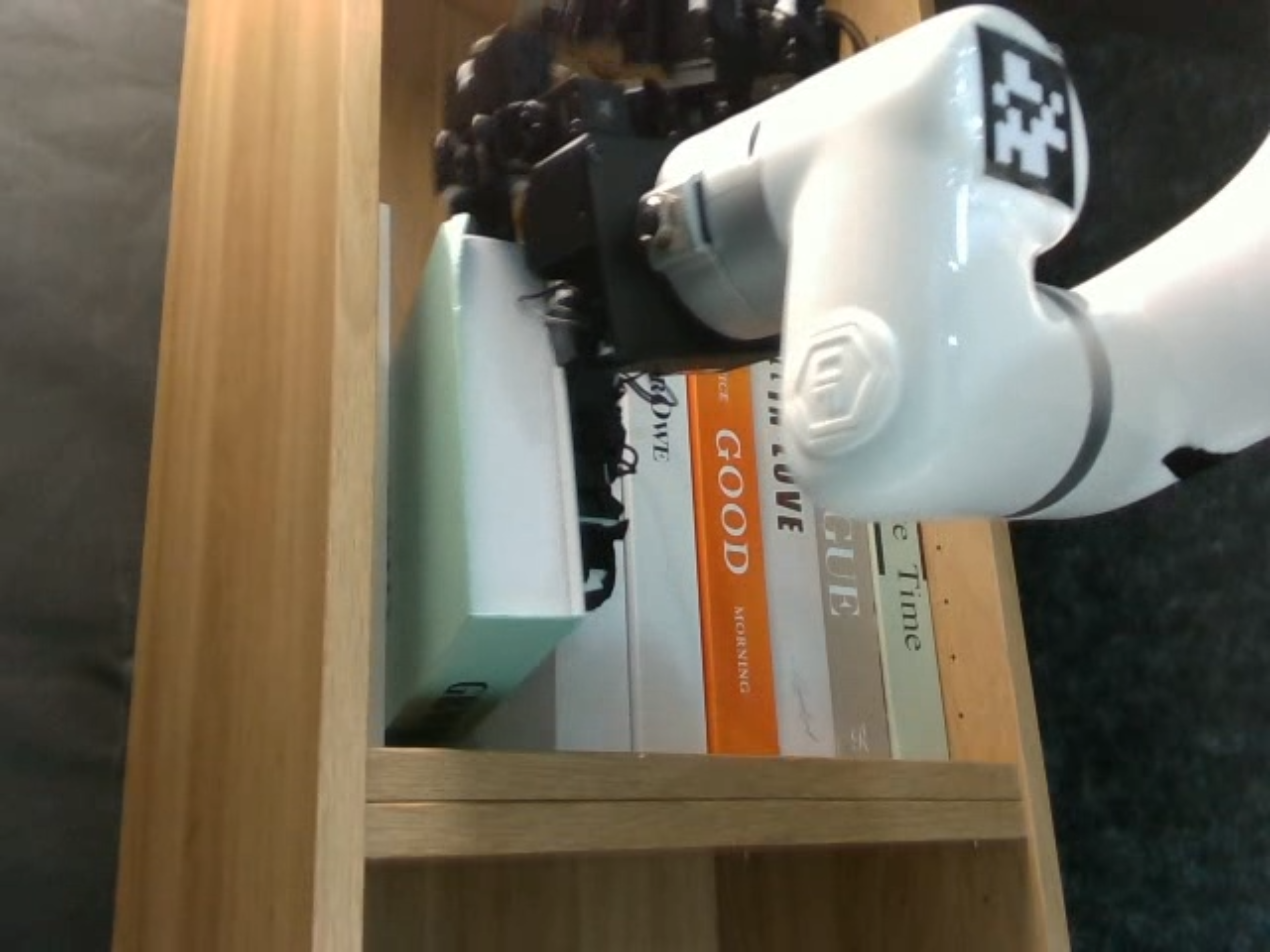}
{\footnotesize (b) Failed grasp}
\end{minipage}
\caption{\textbf{Real-World Failure Cases.} Manipulating a non-target object (left) and failing to stabilize the target after singulation (right).}
\label{app:fig:failures}
\end{figure}

\para{Failure Cases} Figure~\ref{app:fig:failures} shows two failures in real-world singulation: manipulating a non-target object and failing to catch and stabilize the exposed target. The latter can occur after successful exposure, preventing transport to the goal.

\makeatletter
\patchcmd{\BRorg@thebibliography}{\section*{References}}{\section*{Appendix References}}{}{\PackageError{ContactExplorer}{Appendix bibliography heading patch failed}{}}
\patchcmd{\BRorg@thebibliography}{\addcontentsline{toc}{section}{References}}{\addcontentsline{toc}{section}{Appendix References}}{}{\PackageError{ContactExplorer}{Appendix bibliography contents patch failed}{}}
\makeatother
\putbib[citation,appendix-bibliography]
\end{bibunit}

\end{document}